\documentclass{article}

\PassOptionsToPackage{numbers,compress}{natbib}

\usepackage[preprint]{neurips_2026}
\usepackage[utf8]{inputenc}
\usepackage[T1]{fontenc}

\usepackage{microtype}
\usepackage{graphicx}
\usepackage{booktabs}
\usepackage{multirow}
\usepackage[table]{xcolor}
\usepackage{tabularx}
\usepackage{makecell}
\usepackage{hyperref}
\usepackage{amsmath}
\usepackage{amssymb}
\usepackage{mathtools}
\usepackage{pifont}
\usepackage{float}
\usepackage{xcolor}
\usepackage{fontawesome5}
\usepackage{wrapfig}
\usepackage{etoc}

\newcommand{\ie}{\textit{i}.\textit{e}.}
\newcommand{\eg}{\textit{e}.\textit{g}.}
\newcommand{\cf}{\textit{cf.}}

\newcommand{\good}{\textcolor{green!60!black}{\faSmile}}
\newcommand{\neutral}{\textcolor{yellow!70!black}{\faMeh}}
\newcommand{\bad}{\textcolor{red!75!black}{\faFrown}}

\usepackage{color}
\definecolor{citecolor}{RGB}{66,168,235}
\definecolor{linkcolor}{RGB}{255,0,0}
\hypersetup{colorlinks=true,citecolor=citecolor,linkcolor=linkcolor}

\newcolumntype{C}{>{\centering\arraybackslash}X}

\title{SwiftI2V: Efficient High-Resolution Image-to-Video Generation via Conditional Segment-wise Generation}

\author{%
  YaoYang Liu \\
  HKUST \\
  \texttt{yliurj@connect.ust.hk} \\
  \And
  Yuechen Zhang \\
  CUHK \\
  \texttt{zhangyc@link.cuhk.edu.hk} \\
  \And
  Wenbo Li \\
  Joy Future Academy \\
  \texttt{fenglinglwb@gmail.com} \\
  \And
  Yufei Zhao \\
  HKU \\
  \texttt{zhaoyufei@connect.hku.hk} \\
  \And
  Rui Liu \\
  HUAWEI Research \\
  \texttt{ruiliu011@gmail.com} \\
  \And
  Long Chen\thanks{Corresponding author.} \\
  HKUST \\
  \texttt{longchen@ust.hk} \\
}

\begin{document}

\maketitle
\begin{abstract}
High-resolution image-to-video (I2V) generation aims to synthesize realistic temporal dynamics while preserving fine-grained appearance details of the input image. At 2K resolution, it becomes extremely challenging, and existing solutions suffer from various weaknesses: 1) end-to-end models are often prohibitively expensive in memory and latency; 2) cascading low-resolution generation with a generic video super-resolution tends to hallucinate details and drift from input-specific local structures, since the super-resolution stage is not explicitly conditioned on the input image.
To this end, we propose \textbf{SwiftI2V}, an efficient framework tailored for high-resolution I2V. Following the widely used two-stage design, it addresses the efficiency--fidelity dilemma by first generating a low-resolution motion reference to reduce token costs and ease the modeling burden, then performing a strongly image-conditioned 2K synthesis guided by the motion to recover input-faithful details with controlled overhead. Specifically, to make generation more scalable, SwiftI2V introduces \emph{Conditional Segment-wise Generation (CSG)} to synthesize videos segment-by-segment with a bounded per-step token budget, and adopts \emph{bidirectional contextual interaction} within each segment to improve cross-segment coherence and input fidelity. On VBench-I2V at 2K resolution, SwiftI2V achieves performance comparable to end-to-end baselines while reducing total GPU-time by \(202\times\). Particularly, it enables practical 2K I2V generation on a single datacenter GPU (\eg, H800) or consumer GPU (\eg, RTX 4090).
\end{abstract}

\begin{figure*}[t]
    \centering
    \includegraphics[width=\textwidth]{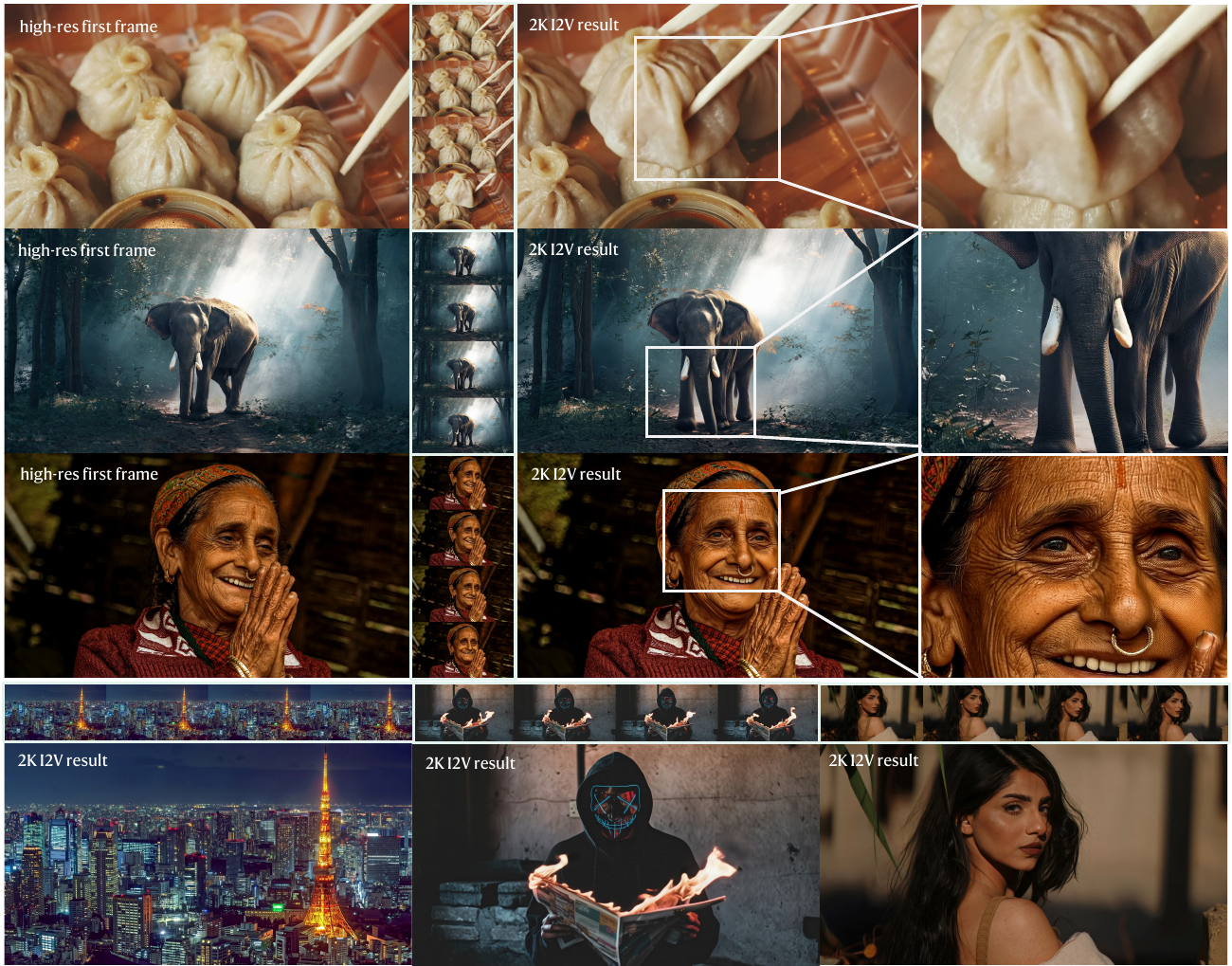}
    \vspace{-1em}
    \caption{Given a high-quality first frame, SwiftI2V enables faithful 2K image-to-video generation with fine-grained details in 2 minutes.}
    \label{fig:teaser}
\end{figure*}

\section{Introduction}

Recent advances in Diffusion Transformer (DiT)~\cite{dit} architectures have steadily improved the perceptual quality and temporal coherence of video generation~\cite{wan,hunyuan1.5,cogvideox,ltx2}. To achieve higher-quality video generation, high-resolution synthesis (\eg, 2K and above) has become an increasingly important direction. Most existing studies focus on text-to-video (T2V), enabling models to produce high-resolution dynamic content aligned with textual semantics~\cite{flashvideo,turbo2k,histream,cinescale}. However, in many real-world applications, users already have a high-resolution image and wish to \textbf{generate a plausible dynamic video while faithfully preserving the image's spatial structure and fine-grained textures}, \ie, image-to-video (I2V).
Despite extensive progress in high-resolution T2V~\cite{flashvideo,turbo2k,histream,cinescale}, efficient 2K-scale I2V with strong image conditioning remains challenging in practice.

There are two challenges for high-resolution I2V. The first is \textit{computational scaling at high resolution}. The number of visual tokens grows rapidly with spatial resolution, making attention-based generation expensive in computation and memory. The second is \textit{fidelity under strong image conditioning}, which is particularly stringent for I2V. The goal is not only to generate plausible motion for the input image, but also to preserve input-specific high-frequency details (\eg, textures, identity cues) with minimal drift across frames. At higher resolutions, tolerance for appearance drift becomes even smaller.

Currently, research on high-resolution I2V remains relatively limited, and two practical paradigms are commonly considered:
1) \textbf{End-to-end}: End-to-end high-resolution generation with a single model~\cite{cinescale} is conceptually simple and can sometimes yield high-fidelity outputs, but must \textit{process all tokens while jointly learning global motion and fine details.}
Such a coupled learning objective often necessitates a larger backbone and more sampling steps. Meanwhile, processing all tokens drives GPU memory usage and computation to prohibitive levels, making training and inference difficult to scale.
2) \textbf{LR+VSR}: One can first generate a low-resolution (LR) video to reduce spatiotemporal modeling cost, and then upscale it using a relatively small video super-resolution (VSR) model~\cite{diffvsr,streamdiffvsr,seedvr,turbovsr}.
This improves efficiency, but the VSR stage is often \textit{not explicitly guided by the input image}, making it prone to hallucinated details and input-structure drift.
% While this pipeline can be more efficient, the VSR stage is often \textit{not explicitly guided by the input image}, \ie, overlooking strong I2V constraints. Consequently, it may hallucinate details and drift from input-specific structures and textures under strong I2V conditioning.

\begin{wraptable}[8]{r}{0.48\textwidth}
\centering
\vspace{-0.6em}
\footnotesize
\setlength{\tabcolsep}{7pt}
\begin{tabular}{lccc}
\toprule
Pipeline & Runtime & Memory & I2V fidelity \\
\midrule
End-to-end & \bad & \bad & \good \\
LR + VSR & \good & \neutral & \bad \\
SwiftI2V & \good & \good & \good \\
\bottomrule
\end{tabular}
\caption{\textbf{Qualitative comparison} of common 2K I2V pipelines under strong image condition.}
\label{tab:intro_tradeoff}
\end{wraptable}

Despite substantial progress, existing high-resolution I2V pipelines still struggle to balance \emph{efficiency} and \emph{fidelity}.
To address both challenges simultaneously, we propose \textbf{SwiftI2V}, an efficient framework tailored for 2K-resolution I2V, as shown in Table~\ref{tab:intro_tradeoff}. SwiftI2V balances efficiency and fidelity by starting with low-resolution motion generation to reduce token costs, and then proceeding to a 2K refinement stage that simultaneously controls computational overhead and introduces strong image conditioning for detail synthesis.

Our key observation is that globally coherent motion can be reliably inferred at much lower spatial resolution, whereas preserving input-specific high-frequency structures is primarily a high-resolution refinement problem that hinges on \emph{strong conditioning on the given image.} This observation naturally fits a motion--detail decoupled two-stage design that is widely adopted in recent high-resolution video generation~\cite{flashvideo,turbo2k}, where a low-resolution stage handles motion and a high-resolution stage handles appearance. SwiftI2V follows this common framework, and focuses our design on \emph{how each stage is realized for 2K I2V}: the low-resolution stage focuses on global motion and coarse appearance, while the high-resolution stage is cast as a \textit{conditional high-resolution video generator} that natively synthesizes 2K frames under joint image and motion conditioning, rather than a generic video super-resolution model. To close the train--test gap at the stage interface, we employ a simple stage-transition strategy that produces Stage~I-like degraded LR videos for training Stage~II, enabling it to handle low-resolution generation artifacts that generic VSR cannot address.

For scalability, we further introduce \textbf{Conditional Segment-wise Generation (CSG)}, which partitions the temporal sequence into bounded segments for controllable memory and streaming generation.
Within each segment, an image-anchored \textbf{bidirectional contextual interaction} lets neighboring and current segments interact, mitigating discontinuities and error accumulation while improving fidelity.
% To improve scalability, we further introduce \textbf{Conditional Segment-wise Generation (CSG)}, which partitions the temporal sequence into segments with bounded per-step token budgets, enabling controllable memory and efficient streaming generation. 
% Within each segment, we introduce the image-anchored \textbf{bidirectional contextual interaction}, which injects input-image information and enables neighboring and current segments to interact bidirectionally, mitigating cross-segment discontinuities and error accumulation while improving quality and fidelity.

Our \textbf{contributions} are summarized as follows:\par\noindent
\textbf{(i)} We propose \textbf{SwiftI2V}, an efficient high-resolution I2V framework that tackles the efficiency--fidelity dilemma. On VBench-I2V at 2K, SwiftI2V matches strong end-to-end high-resolution baseline~\cite{cinescale} on key I2V metrics while reducing total GPU-time by \(202\times\), and supports practical 2K I2V on a single consumer GPU (\eg, RTX~4090).
\par\noindent
\textbf{(ii)} We propose \textbf{Conditional Segment-wise Generation (CSG)} with bidirectional contextual interaction, bounding the per-step 2K token budget for segment-wise streaming while avoiding autoregressive error accumulation.
\par\noindent
\textbf{(iii)} We introduce a simple \textbf{stage-transition training} strategy that injects Stage~I-like artifacts into Stage~II inputs, reducing the cascade's train--test gap.

\section{Related Work}

\noindent\textbf{Video Diffusion Models (VDMs).}
VDMs first introduced diffusion models to video generation~\cite{vdm}. Subsequent works adopted latent diffusion models (LDMs)~\cite{ldm}, performing diffusion in compressed latent spaces for better scalability~\cite{vldm}. Recent VDMs further incorporate Transformer~\cite{dit} architectures, forming the dominant modeling paradigm and exhibiting strong generative capacity in terms of visual fidelity, aesthetic quality, and spatiotemporal coherence~\cite{cogvideox,wan,hunyuan1.5}.

From a task perspective, most existing VDMs focus on text-to-video (T2V) generation~\cite{cogvideox,wan,hunyuan1.5}. Unlike T2V, which relies only on text, image-to-video (I2V) uses an input image as a strong condition and requires strict spatial and semantic consistency over time. Thus, I2V differs from T2V in objectives and difficulty. Some works adapt T2V models to I2V by introducing image conditions~\cite{wan,hunyuan1.5}, while others are designed for I2V~\cite{motioni2v,i2vgen}.

\noindent\textbf{High-Resolution Video Generation.}
It is an important research direction due to the increased demand for fine-grained visual details and spatiotemporal consistency. In the T2V task, prior works explicitly investigate scaling diffusion models to high resolutions through high-resolution training~\cite{guo2024make,flashvideo,turbo2k,turbovsr} or tuning-free~\cite{cinescale} strategies; in the I2V task, recent studies have also begun to explore high-resolution generation under strong image conditions~\cite{cinescale}. However, these approaches typically incur substantially increased computational and memory costs, which limits their scalability to higher resolutions or more constrained settings.
A common alternative is to generate videos at low resolution and apply video super-resolution as a post-processing step~\cite{upscaleavideo,diffvsr,streamdiffvsr,star,seedvr}, but such two-stage pipelines often struggle to recover faithful fine details. For I2V, these methods may compromise input image fidelity.

\noindent\textbf{Efficient Video Generation.}
The multi-step iterative inference process of diffusion models, together with the quadratic complexity of attention mechanisms, poses significant challenges to efficient video generation for DiT models. To address this issue, a large body of work proposes efficiency-oriented techniques, such as reducing denoising steps via distillation~\cite{wang2023videolcm,casuvid} and accelerating attention computation through causal modeling~\cite{casuvid,magi-1,histream} or related optimizations. These methods have also been incorporated into high-resolution T2V generation~\cite{histream}. However, for high-resolution I2V, the applicability of existing acceleration methods remains insufficiently explored. Concurrent work LTX-2~\cite{ltx2} is an efficient joint audio--visual foundation model supporting 2K I2V, but it is not tailored to this strongly image-anchored setting, leaving a fidelity--motion gap.

\begin{figure*}[t]
    \centering
    \includegraphics[width=\textwidth]{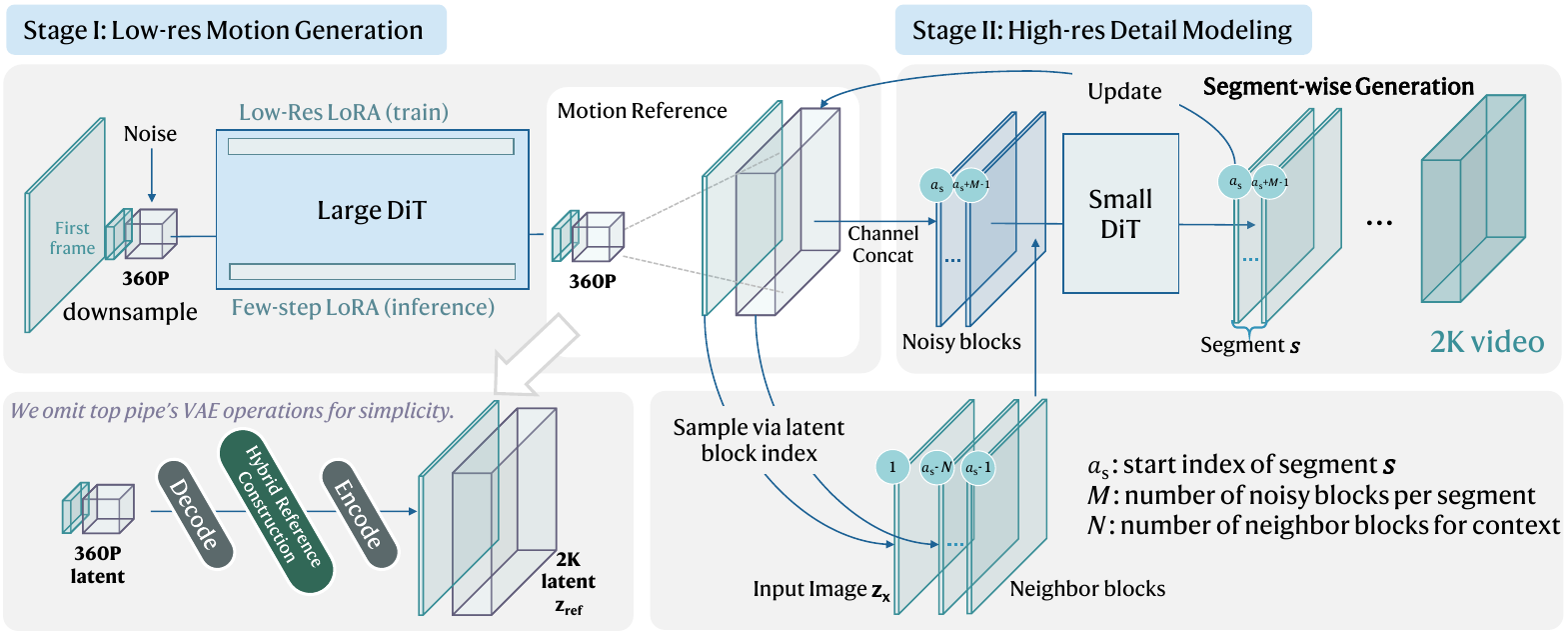}
    
    \vspace{-1em}
    \caption{\textbf{Overview of SwiftI2V.}
    Stage~I generates a low-resolution motion reference, which is fused with the input high-resolution image and concatenated with Stage~II DiT noise. Stage~II then generates the high-resolution video segment-by-segment.}
    \label{fig:method}
\end{figure*}

\section{Method}

\noindent\textbf{Overview.}
SwiftI2V achieves 2K I2V within a tractable budget via two stages (Figure~\ref{fig:method}): Stage~I generates a low-resolution motion reference, and Stage~II synthesizes input-faithful high-resolution details through a lightweight conditioning interface. Stage~II further uses CSG with bidirectional contextual interaction to control the per-step token budget while preserving fidelity.
% SwiftI2V is an efficient framework that achieves high-quality 2K I2V generation within a tractable computational budget. Figure~\ref{fig:method} shows an overview of our framework.
% SwiftI2V has two stages: Stage~I generates a low-resolution video to capture globally coherent and plausible motion patterns, and Stage~II synthesizes input-faithful details and generates a final high-resolution result conditioned on the motion guidance via a lightweight conditioning interface.
% In Stage~II, we propose Conditional Segment-wise Generation (CSG) with bidirectional contextual interaction, which keeps fidelity while controlling the per-step token budget.

\subsection{Two-stage High-Resolution I2V Framework}
\label{sec:2stage}

Given a high-resolution input image \(\mathbf{x}\in\mathbb{R}^{H\times W\times 3}\),
our goal is to synthesize a \(T\)-frame 2K video
\(\hat{\mathbf{V}}\in\mathbb{R}^{T\times H\times W\times 3}\)
that exhibits realistic temporal dynamics while faithfully preserving input-specific spatial structure and fine-grained textures.
Below we describe how each stage is instantiated and how motion reference is transferred across stages.

\noindent\textbf{Stage I: Low-Resolution Motion Reference Generation.}
Stage~I models globally coherent motion at low resolution by downsampling the input image as \(\mathbf{x}^{\mathrm{LR}}=\mathrm{Down}(\mathbf{x})\)
and using a large-capacity DiT backbone \(\mathcal{G}_1\) to generate a low-resolution video \(\hat{\mathbf{V}}^{\mathrm{LR}}\)
as a motion and structure reference:
\begin{equation}
\hat{\mathbf{V}}^{\mathrm{LR}}=\mathcal{G}_1\!\left(\mathbf{x}^{\mathrm{LR}}\right)
\in\mathbb{R}^{T\times H_{\mathrm{LR}}\times W_{\mathrm{LR}}\times 3}.
\label{eq:stage1_generate}
\end{equation}
Operating at low resolutions greatly reduces the token count, allowing us to afford a large-capacity backbone that robustly learns motion priors while keeping the compute budget manageable.
On top of this backbone, we train a \textit{Low-Res LoRA}~\cite{lora} for resolution adaptation, and further couple it with an off-the-shelf \textit{Few-Step LoRA}~\cite{lightx2v} at inference to reduce the number of denoising steps, yielding a fast yet motion-faithful reference generator.

\noindent\textbf{Pixel-Space Transition: Hybrid Reference Construction.}
To transfer Stage~I motion priors to high-resolution synthesis, we upsample its output to the target resolution:
\begin{equation}
\mathbf{V}^{\mathrm{up}}=\mathrm{Up}\!\left(\hat{\mathbf{V}}^{\mathrm{LR}}\right)
\in\mathbb{R}^{T\times H\times W\times 3}.
\label{eq:upsample}
\end{equation}
Let \(\mathbf{V}^{\mathrm{up}}_\tau\in\mathbb{R}^{H\times W\times 3}\) denote the \(\tau\)-th frame,
\(\tau\in\{1,\ldots,T\}\).
We then construct a hybrid reference video \(\tilde{\mathbf{V}}\) by replacing the first frame with the input:
\begin{equation}
\tilde{\mathbf{V}}_\tau=
\begin{cases}
\mathbf{x}, & \tau=1,\\
\mathbf{V}^{\mathrm{up}}_\tau, & \tau=2,\ldots,T.
\end{cases}
\label{eq:first_frame_replace}
\end{equation}
This first-frame replacement injects the input image as a boundary condition to reduce drift and first-frame mismatch compared with traditional VSR pipelines, while frames \(\mathbf{V}^{\mathrm{up}}_{2:T}\) preserve Stage~I motion and structure as a stable reference for Stage~II.
% This first-frame replacement explicitly injects the input image as a boundary condition, mitigating drift and first-frame mismatch compared with traditional VSR pipelines, thereby improving I2V fidelity.
% Meanwhile, frames \(\mathbf{V}^{\mathrm{up}}_{2:T}\) preserve the motion tendency and structure generated in Stage~I, providing a stable reference for Stage~II.

\noindent\textbf{Stage II: High-Resolution Video Synthesis.}
Stage~II focuses on synthesizing input-faithful high-frequency details conditioned on the Stage~I motion reference and the input appearance constraint.
Since it does not need to re-model motion from scratch, a smaller DiT backbone is sufficient, allowing its limited capacity to be devoted to high-frequency detail synthesis rather than motion modeling. To further reduce the number of tokens at high resolution, Stage~II adopts a 3D VAE with higher downsampling factors \((16,16,4)\)~\cite{wan}; Appendix~\ref{app:vae-fidelity} validates its 2K reconstruction fidelity.
Let \(\mathcal{E}^{\mathrm{HR}}\) and \(\mathcal{D}^{\mathrm{HR}}\) denote the VAE encoder and decoder.
We encode the hybrid reference video \(\tilde{\mathbf{V}}\) and the input image \(\mathbf{x}\) as
\begin{equation}
\mathbf{z}_{\mathrm{ref}} = \mathcal{E}^{\mathrm{HR}}(\tilde{\mathbf{V}}),
\qquad
\mathbf{z}_{x} = \mathcal{E}^{\mathrm{HR}}(\mathbf{x}),
\label{eq:encode_ref_x}
\end{equation}
where \(\mathbf{z}_{\mathrm{ref}}\in\mathbb{R}^{t\times h\times w\times c}\) and
\(\mathbf{z}_{x}\in\mathbb{R}^{h\times w\times c}\).
Here \((t,h,w)\) are the latent spatiotemporal dimensions.

During denoising step \(k\), let \(\mathbf{z}_k\in\mathbb{R}^{t\times h\times w\times c}\) be the noisy latent,
and write it along the temporal axis as \(t\) blocks \(\mathbf{z}_k=(\mathbf{z}_{k,1},\ldots,\mathbf{z}_{k,t})\),
\(\mathbf{z}_{k,i}\in\mathbb{R}^{h\times w\times c}\).
Since the 3D VAE encodes the first frame separately during encoding, we further anchor the high-resolution appearance information by replacing the first block of the noisy latent \(\mathbf{z}_k\) with \(\mathbf{z}_x\):
\begin{equation}
\bar{\mathbf{z}}_{k,i}=
\begin{cases}
\mathbf{z}_{x}, & i=1,\\
\mathbf{z}_{k,i}, & \text{otherwise}.
\end{cases}
\label{eq:replace_first_slice}
\end{equation}
and concatenate \(\bar{\mathbf{z}}_{k}\) with \(\mathbf{z}_{\mathrm{ref}}\) along the channel dimension
to construct the Stage~II DiT input:
\begin{equation}
\mathbf{u}_k
=
\operatorname{Concat}_{c}\!\left(\bar{\mathbf{z}}_{k},\,\mathbf{z}_{\mathrm{ref}}\right)
\in \mathbb{R}^{t\times h\times w\times 2c}.
\label{eq:concat_input_stage2}
\end{equation}
Here, \(\mathbf{z}_{x}\) acts as an explicit appearance anchor, while \(\mathbf{z}_{\mathrm{ref}}\) provides motion cues and structural appearance information.
We then denoise \(\mathbf{z}_k\) to obtain \(\mathbf{z}_0\) in combination with our Conditional Segment-wise strategy.
Finally, the high-resolution video is decoded as:
$\hat{\mathbf{V}} = \mathcal{D}^{\mathrm{HR}}(\mathbf{z}_0).$

\subsection{Conditional Segment-wise Generation (CSG)}
\label{sec:csg}

\begin{wrapfigure}[12]{r}{0.5\textwidth}
    \centering
    \vspace{-1.5em}
    \includegraphics[width=\linewidth]{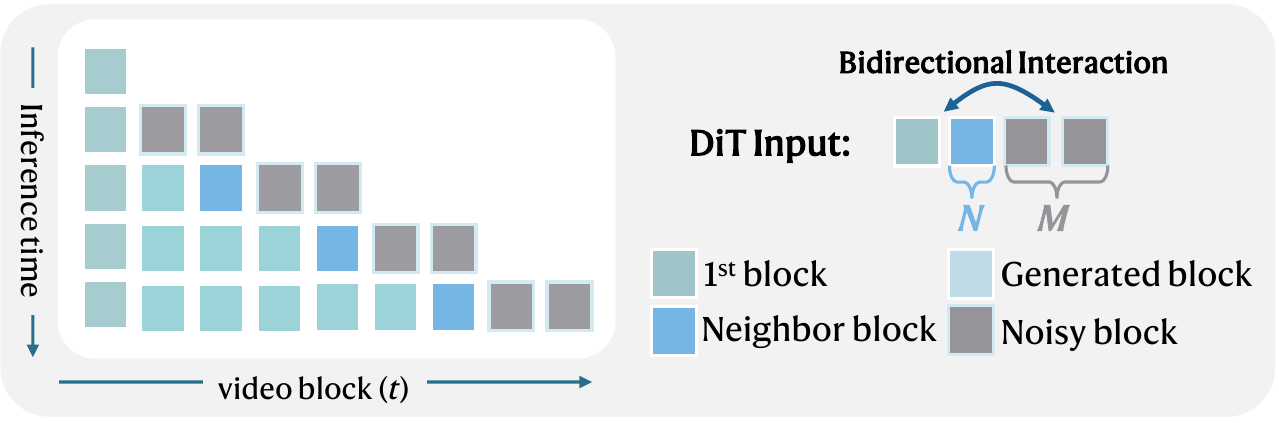}
    \vspace{-1em}
    \caption{\textbf{Conditional Segment-wise Generation of SwiftI2V.} SwiftI2V adopts a CSG strategy in Stage II. To ensure fidelity and mitigate error accumulation, SwiftI2V allows bidirectional interaction between conditioning blocks and noisy blocks.}
    \label{fig:CSG}
\end{wrapfigure}

Even with a highly compressed VAE, 2K Stage~II still has many visual tokens. Since the input image and Stage~I reference already provide global structure and dynamics, Stage~II mainly needs to recover high-frequency details with smooth temporal transitions. We therefore introduce CSG, which denoises high-resolution latents in short temporal segments with bounded per-step token budgets. The term \emph{conditional} emphasizes native 2K synthesis under the input-image anchor and Stage~I motion reference, rather than low-resolution upsampling.

% Although Stage~II uses a VAE with a higher compression ratio, the number of visual tokens at 2K resolution remains substantial. Moreover, Stage~II operates in a refinement setting: the input image and the low-resolution motion reference from Stage~I already provide strong global structure and dynamics, so the remaining challenge is to recover high-frequency details while ensuring smooth transitions over time. Motivated by these observations, we introduce CSG, which denoises the high-resolution latents in short temporal segments with a bounded per-step token budget. The term \emph{conditional} emphasizes that each segment is synthesized natively at high resolution under the joint conditioning of the input-image anchor and the Stage~I motion reference, rather than being upsampled from low-resolution outputs.

\noindent\textbf{Temporal Block and Segment-level Windows.}
Following Eq.~\eqref{eq:concat_input_stage2}, the DiT input at diffusion step \(k\) is \(\mathbf{u}_k = \operatorname{Concat}_c(\bar{\mathbf{z}}_k, \mathbf{z}_{\mathrm{ref}})\),
where \(\bar{\mathbf{z}}_k\) is the noised high-resolution latent sequence and \(\mathbf{z}_{\mathrm{ref}}\) is the hybrid reference latent sequence. Split \(\mathbf{u}_k\) along time into \(t\) blocks:
\begin{equation}
\mathbf{u}_k = (\mathbf{u}_{k,1}, \mathbf{u}_{k,2}, \ldots, \mathbf{u}_{k,t}),
\qquad
\mathbf{u}_{k,i} \in \mathbb{R}^{h \times w \times 2c}.
\label{eq:csg_block}
\end{equation}
By the substitutions in Eq.~\eqref{eq:first_frame_replace} and Eq.~\eqref{eq:replace_first_slice}, the first temporal block \(\mathbf{u}_{k,1}\) is anchored to the HR input image, which we denote as the \textit{anchor block}.
Blocks \(\{\mathbf{u}_{k,2},\ldots,\mathbf{u}_{k,t}\}\) are to be generated. CSG aims to inject high-fidelity details consistent with the HR input on top of the motion cues from \(\mathbf{z}_{\mathrm{ref}}\), while keeping the per-step token budget bounded.

We partition the target indices \(\{2,\ldots,t\}\) into \(S\) consecutive, non-overlapping segments, each containing \(M\) noisy blocks. Define
\[
S \triangleq \left\lceil \frac{t-1}{M} \right\rceil,
\qquad
a_s \triangleq 2 + (s-1)M,
\]
and the noisy-block index set of segment \(s\) as
\begin{equation}
\mathcal{I}_s \triangleq \{a_s, a_s+1, \ldots, a_s + M - 1\},
\qquad
s \in \{1,\ldots,S\}.
\label{eq:csg_Is}
\end{equation}
To promote cross-segment continuity, we additionally include a short context consisting of the last \(N\) blocks immediately preceding the current segment. Specifically, we define the neighbor index set
\begin{equation}
\mathcal{N}_s \triangleq
\begin{cases}
\emptyset, & s=1,\\
\{\max(2, a_s - N),\ \ldots,\ a_s - 1\}, & s>1.
\end{cases}
\label{eq:csg_Ns}
\end{equation}
We then construct the segment-wise temporal window
\(\mathcal{W}_s \triangleq \{1\} \cup \mathcal{N}_s \cup \mathcal{I}_s\),
and feed the gathered subsequence into DiT at each diffusion step, as shown in Figure~\ref{fig:CSG}:
\begin{equation}
\mathbf{u}_k^{(s)} \triangleq \big( \mathbf{u}_{k,i} \big)_{i \in \mathcal{W}_s}
\in \mathbb{R}^{|\mathcal{W}_s| \times h \times w \times 2c}.
\label{eq:csg_window_input}
\end{equation}
During inference, segments are processed sequentially; once segment \(\mathcal{I}_s\) finishes diffusion, we can decode its frames and cache the last \(N\) blocks as \(\mathcal{N}_{s+1}\), enabling segment-wise streaming output.

\noindent\textbf{Bidirectional Contextual Interaction.}
A key design choice is how conditioning blocks (the \textit{anchor block} and the \textit{neighbor blocks}) are used within the window. Some streaming methods~\cite{histream,casuvid,magi-1,nova} use an auto-regressive (AR) formulation, where previous blocks serve as fixed read-only context for the current blocks. While this controls the token budget, the rigid dependence on imperfect history can cause boundary artifacts and error accumulation in high-fidelity I2V.
% A critical design choice is how conditioning blocks (the \textit{anchor block} and the \textit{neighbor blocks}) are used within the window. In some video generation methods~\cite{histream,casuvid,magi-1,nova}, streaming inference is realized via an auto-regressive (AR) formulation: previously generated blocks are treated as fixed context, and the current blocks get information from this context.
% While this supports streaming and controllable token budget, the ``read-only'' context can cause boundary artifacts and exacerbate error accumulation in high-fidelity I2V due to its rigid dependence on imperfect history.

To better preserve input-image fidelity and to mitigate segment-wise degradation, CSG introduces a bidirectional contextual interaction strategy within each window \(\mathcal{W}_s\): we apply standard attention over \(\mathbf{u}_k^{(s)}\), so that the anchor block, the neighbor blocks, and the current noisy blocks all attend to each other bidirectionally within the window. \textit{As a result, conditioning blocks are not merely static providers of features, but actively participate in the attention computation together with the current noisy blocks.} This lets the context be dynamically reorganized to match the denoising needs of the current segment, facilitating the fusion of anchored HR appearance and reference motion cues and mitigating cascading error accumulation across segments.

Crucially, bidirectional interaction is only used for feature interaction and does not alter previously finalized latents. Although DiT produces predictions for all blocks inside \(\mathcal{W}_s\), we only apply the update to the current segment \(\mathcal{I}_s\) and never write back updates to the conditioning latents:
\begin{equation}
\bar{\mathbf{z}}_{k-1,i} =
\begin{cases}
\mathrm{Update}\!\left(\bar{\mathbf{z}}_{k,i};\ \mathrm{DiT}_\theta(\mathbf{u}_k^{(s)})_i\right), & i \in \mathcal{I}_s,\\
\bar{\mathbf{z}}_{k,i}, & \text{otherwise}.
\end{cases}
\label{eq:csg_update_rule}
\end{equation}

Generated historical segments remain immutable as input, while still providing stronger, more adaptive interactive features inside the attention in the DiT forward pass.

Overall, the design of CSG aligns with the refinement objective in Stage~II---recovering input-faithful details while maintaining smooth temporal transitions. Meanwhile, it bounds per-step high-resolution token budget, improving the model's scalability, and enables segment-wise decoding for low-latency, streaming outputs (Appendix~\ref{app:streaming} provides an exploratory deployment study.)

To mitigate train--test mismatch, we train the Stage~II model using CSG. We use teacher-forcing~\cite{teacherfocing} for the conditioning blocks, which are taken from the ground-truth training video, and compute the diffusion loss only on the noisy blocks indexed by \(\mathcal{I}_s\). This trains the model to exploit the anchored HR appearance and short-range context, on top of the reference guidance \(\mathbf{z}_{\mathrm{ref}}\), to generate high-fidelity segments.

\subsection{Stage-Transition Training}
\label{sec:stage-trans}

Training the two stages separately keeps each stage lightweight and tractable at 2K resolution, but it introduces an interface gap that is common to separately trained cascades: Stage~II is trained with ``clean'' LR inputs (downsampled HR videos), whereas at inference it consumes Stage~I outputs that may contain generation artifacts caused by VAE distortion or low-resolution flickering, leading to error amplification. To close this gap without re-introducing motion modeling into Stage~II, we synthesize Stage~II inputs by lightly corrupting downsampled clips and denoising them with Stage~I.

For a training clip \(V^{HR}\) (with first frame as input image \(x\)), we construct
\[
V^{LR}_{\mathrm{noisy}} \triangleq \mathrm{Down}(V^{HR}) + \sigma \epsilon,\qquad \epsilon \sim \mathcal{N}(0, I),
\]
and \(x^{LR} \triangleq \mathrm{Down}(x)\). Then we denoise it using the Stage~I model,
\begin{equation}
\tilde{V}^{LR}
\triangleq \mathrm{Denoise}_{G_1}\!\left(V^{LR}_{\mathrm{noisy}} \,;\, x^{LR}\right).
\label{eq:stage_transition_denoise}
\end{equation}
The synthesized \(\tilde{V}^{LR}\) preserves the ground-truth motion patterns to the greatest extent while inheriting Stage~I--style artifacts, making it a closer match to Stage~II's inference-time inputs and preserving a reliable motion--appearance supervision signal. We then pair \(\tilde{V}^{LR}\) with the original \(V^{HR}\) and train Stage~II on pairs \((\tilde{V}^{LR}, V^{HR})\). In practice, this simple input synthesis substantially reduces the stage-to-stage gap and enables stable performance when the two separately trained stages are cascaded at inference time. We provide further analysis in Appendix~\ref{app:stage-trans}.

\begin{figure*}[t]
    \centering
    \includegraphics[width=\textwidth]{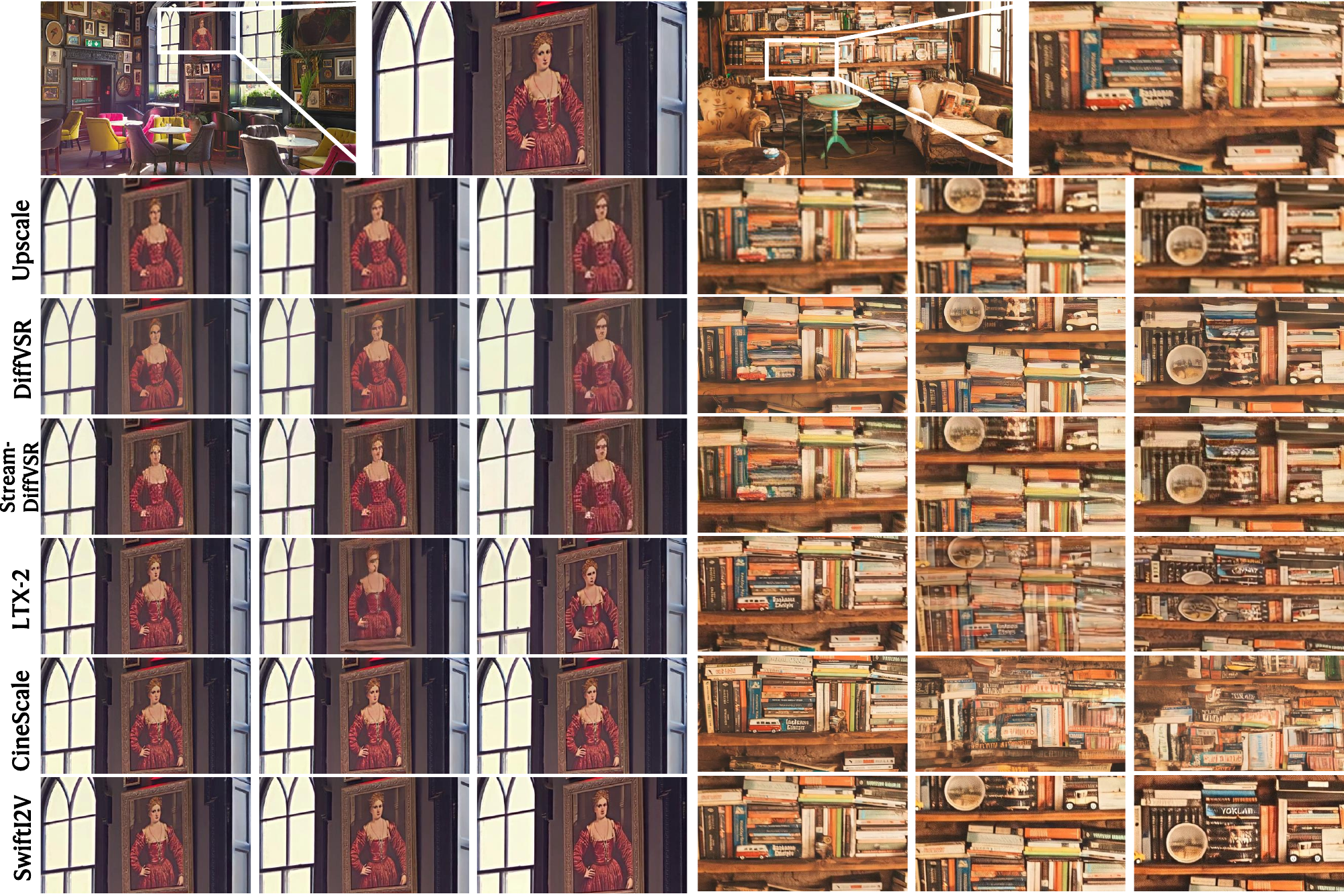}
    \vspace{-1em}
    \caption{Qualitative comparison between SwiftI2V and representative baselines on 2K I2V generation. Best viewed \textbf{zoomed in}.}
    \label{fig:exp}
\end{figure*}

\section{Experiments}
\label{sec:experiments}

\subsection{Experimental Setup}
\label{sec:exp-setup}

Unless stated otherwise, all experiments are conducted on NVIDIA H800 GPUs.
Our default generation setting produces 81-frame videos at 2K resolution (\(2560\times1408\)), where \textbf{Stage~I} generates a low-resolution video at 360P (\(640\times352\)), and \textbf{Stage~II} synthesizes the final 2K result.

\noindent\textbf{Implementation Details.}
For Stage~I, we adopt Wan2.1-I2V-480P~\cite{wan} as the backbone.
We train a LoRA to perform I2V generation at 360P. At inference time, we additionally load an existing few-step LoRA~\cite{lightx2v} to accelerate sampling, enabling 4-step generation in Stage~I.
For Stage~II, we adopt Wan2.2-TI2V-5B~\cite{wan} as the backbone and fully fine-tune its DiT for high-resolution video synthesis. We observe in the experiment that 4-step inference already yields stable and competitive results for this refinement stage, and we therefore use it as the default.
For Conditional Segment-wise Generation (CSG), we use \(M{=}3\) and \(N{=}1\) for inference; see Appendix~\ref{app:mn-selection} for the experiment that motivates this choice.
Since 2K training data is relatively limited, we employ a curriculum strategy: we first train with 1080P videos from OpenViD-HD~\cite{openvid}, then continue for 10K steps with 90K 2K videos from UltraVideo~\cite{ultravideo}, mixed with our synthesized samples.

\noindent\textbf{Evaluation Metrics.}
We use VBench-I2V~\cite{vbench++} as our primary evaluation suite. It measures I2V-specific fidelity (\eg, \textit{i2v subject} and \textit{i2v background}) as well as general video quality metrics. I2V generation is conditioned on both the input image and text, where the text prompts are taken from the official VBench-I2V prompt set. We also report runtime and GPU memory efficiency in Section~\ref{sec:efficiency-analysis} to validate different pipelines' practicality.

\begin{table*}[t]
\centering
\small
\setlength{\tabcolsep}{5pt}
\renewcommand{\arraystretch}{1.12}
\caption{\textbf{Main results on VBench-I2V at 2K resolution.}
All metrics are higher-is-better (\(\uparrow\)).
\textbf{Bold} and \underline{underline} indicate the best and second-best results, respectively. 
% \textcolor[gray]{0.35}{Upscale} (Stage~I output upscaled) serves as a reference baseline and is not included in ranking.
}
\label{tab:result}
\begin{tabularx}{\textwidth}{@{}l*{6}{>{\centering\arraybackslash}X}@{}}
\toprule
\textbf{Model} &
\shortstack[c]{\textbf{Total} \\ \textbf{Score}\(\uparrow\)} &
\shortstack[c]{\textbf{I2V} \\ \textbf{Subject}\(\uparrow\)} &
\shortstack[c]{\textbf{I2V} \\ \textbf{Background}\(\uparrow\)} &
\shortstack[c]{\textbf{Dynamic} \\ \textbf{Degree}\(\uparrow\)} &
\shortstack[c]{\textbf{Aesthetic} \\ \textbf{Quality}\(\uparrow\)} &
\shortstack[c]{\textbf{Motion} \\ \textbf{Smoothness}\(\uparrow\)} \\
\midrule
\textcolor[gray]{0.35}{Upscale} & \textcolor[gray]{0.35}{6.4173} & \textcolor[gray]{0.35}{0.9881} & \textcolor[gray]{0.35}{0.9946} & \textcolor[gray]{0.35}{0.2805} & \textcolor[gray]{0.35}{0.6402} & \textcolor[gray]{0.35}{0.9910} \\
\midrule
DiffVSR~\cite{diffvsr} &
6.4228 & 0.9878 & 0.9907 & 0.2917 & \textbf{0.6567} & 0.9890 \\
Stream-DiffVSR~\cite{streamdiffvsr} &
\underline{6.4240} & 0.9879 & 0.9933 & \textbf{0.3374} & 0.6447 & 0.9856 \\
LTX--2~\cite{ltx2} &
6.3579 & \underline{0.9914} & 0.9932 & 0.0488 & \underline{0.6534} & \textbf{0.9939} \\
CineScale\textcolor{cyan}{$^{\dagger}$}~\cite{cinescale} &
6.3638 & \textbf{0.9924} & \underline{0.9973} & 0.1667 & 0.6462 & \underline{0.9909} \\
\midrule
SwiftI2V (ours) &
\textbf{6.4244} & 0.9910 & \textbf{0.9975} & \underline{0.3008} & 0.6496 & 0.9885 \\
\bottomrule
\end{tabularx}

\raggedright\footnotesize
\textcolor{cyan}{$^{\dagger}$} Tested on a random subset due to high computational cost. Appendix~\ref{app:cinescale-subset} shows the result of SwiftI2V on this subset.
\end{table*}

\subsection{Comparison with High-Resolution I2V Methods}
\label{sec:main-comparison}

\noindent\textbf{Baselines.}
We compared SwiftI2V against representative 2K I2V pipelines, including CineScale~\cite{cinescale}, an end-to-end method that directly generates 2K videos within a single model; LTX-2~\cite{ltx2}, an efficient audio--visual foundation model that also supports 2K I2V. We also compare SwiftI2V with two VSR-based pipelines, DiffVSR~\cite{diffvsr} and Stream-DiffVSR~\cite{streamdiffvsr}. For the VSR baselines, to control for the low-resolution generation input, we feed them the same Stage~I outputs from SwiftI2V and upscale the resulting videos to 2K.

\noindent\textbf{Results.}
Table~\ref{tab:result} summarizes key 2K VBench-I2V~\cite{vbench++} results; full results are in Appendix~\ref{app:full-results}. SwiftI2V achieves the best total score (6.4244) and I2V Background score (0.9975), while keeping competitive I2V Subject fidelity (0.9910) and aesthetic quality. VSR pipelines show weaker I2V faithfulness, especially background consistency, indicating that post-hoc super-resolution is less reliable at recovering input-specific structures under strong image conditioning.
CineScale and LTX-2 obtain lower Dynamic Degree under HR image conditioning (0.1667 and 0.0488), which may contribute to their stronger scores on some metrics that favor visual quality and temporal stability; Stream-DiffVSR's higher Dynamic Degree largely reflects flickering rather than coherent motion (Appendix~\ref{app:dynamic-degree}). In addition, the \textcolor[gray]{0.35}{Upscale} row shows that Stage~I already provides a strong motion reference (Dynamic Degree 0.2805), allowing Stage~II to focus on 2K detail synthesis rather than correcting severe motion failures from scratch.
% Table~\ref{tab:result} summarizes key VBench-I2V~\cite{vbench++} results at 2K resolution, covering I2V faithfulness, perceptual quality, and motion-related criteria (the complete evaluation results are provided in Appendix~\ref{app:full-results}). SwiftI2V achieves the highest total score of 6.4244 and attains the highest I2V Background score (0.9975), while maintaining competitive I2V Subject fidelity (0.9910) and aesthetic quality. In comparison, VSR-based pipelines exhibit reduced I2V faithfulness in this setting, with a particularly noticeable drop on background consistency, suggesting that post-hoc super-resolution may be less reliable at recovering input-specific structures and textures under strong image conditioning.
% We also observe that CineScale and LTX-2 obtain lower Dynamic Degree scores under HR image conditioning (0.1667 and 0.0488, respectively); this may contribute to their stronger scores on some metrics that favor visual quality and temporal stability. While Stream-DiffVSR reports a higher Dynamic Degree than SwiftI2V, this largely reflects temporal flickering rather than coherent motion; a detailed analysis is provided in Appendix~\ref{app:dynamic-degree}. In addition, the \textcolor[gray]{0.35}{Upscale} row indicates that our Stage~I already produces videos with a high Dynamic Degree (0.2805), providing a high-quality motion reference for Stage~II. This also makes Stage~II focus on easier high-resolution detail synthesis rather than correcting severe motion failures from scratch.

As shown in Figure~\ref{fig:exp}, SwiftI2V preserves higher fidelity and finer details than VSR-based pipelines, while the strong generative capacity of Stage~I enables plausible and coherent dynamics.

\subsection{Efficiency and Memory Analysis}
\label{sec:efficiency-analysis}

\noindent\textbf{Settings.}
We further evaluated inference efficiency and memory consumption under our experiment settings. Table~\ref{tab:efficiency} reports the \textit{total wall-clock time} to obtain the final 81-frame 2K video. For VSR pipelines (DiffVSR and Stream-DiffVSR), the reported time includes both the base video generation stage (our Stage~I) and the subsequent super-resolution stage.

\noindent\textbf{Results.}
As shown in Table~\ref{tab:efficiency}, SwiftI2V achieves the \textbf{lowest latency}, taking \(111\)s on a single GPU. CineScale requires \(4\) GPUs and \(5600\)s; measured in GPU-time (\(\#\mathrm{GPUs}\times\mathrm{time}\)), this corresponds to \(111\) vs.\ \(22400\) GPU\(\cdot\)s, \ie, \textbf{a \(202\times\) reduction}, demonstrating the efficiency advantage of our two-stage design at 2K resolution. SwiftI2V is also faster than LTX-2, a strong efficiency-oriented baseline, on a single GPU (111s vs.\ 152s, \ie, \(1.37\times\) speedup).

Table~\ref{tab:efficiency} further reports the \textit{peak GPU memory} measured over the entire generation pipeline. Without any further memory optimizations like model off-loading or quantization, SwiftI2V has the lowest peak memory usage during inference, which is \(33.5\)GB and occurs in Stage~I, while Stage~II peaks at only \(20.8\)GB. This indicates that CSG effectively controls the memory consumption of the 2K generation stage. Moreover, by incorporating existing memory-saving inference techniques~\cite{diffsynth_studio} for Stage~I, SwiftI2V can be deployed on a single consumer-grade GPU (\eg, RTX~4090); detailed configurations and results are provided in Appendix~\ref{app:4090}. This substantially lowers the hardware barrier for practical 2K I2V generation.

\begin{table}[t]
\centering
\footnotesize
\begin{minipage}[t]{0.44\textwidth}
\centering
\caption{\textbf{Efficiency and peak memory} for generating an 81-frame 2K video (\(2560\times1408\)) on NVIDIA H800 GPUs.}
\label{tab:efficiency}
\begin{tabular*}{\linewidth}{@{\extracolsep{\fill}}lccc}
\toprule
Model & \#GPUs & Peak mem. & Time \\
\midrule
DiffVSR & 1 & 58.1GB & 3386s \\
Stream-DiffVSR & 1 & 49.2GB & 141s \\
LTX-2 & 1 & 45.7GB & 152s \\
CineScale & 4 & 42.7GB & 5600s \\
\textbf{SwiftI2V} & \textbf{1} & \textbf{33.5GB} & \textbf{111s} \\
\bottomrule
\end{tabular*}
\end{minipage}
\hfill
\begin{minipage}[t]{0.53\textwidth}
\centering
\setlength{\tabcolsep}{1.9pt}
\renewcommand{\arraystretch}{1.12}
\caption{\textbf{Ablation results}. All methods share Stage~I; we report Stage~II peak memory and DiT time.}
\label{tab:ablation}
\vspace{0.8\baselineskip}
\begin{tabular}{@{}lcccccc@{}}
\toprule
\raisebox{0.5\height}{Method} &
\shortstack{Stream\\infer.} &
\shortstack{Peak\\mem.\(\downarrow\)} &
\shortstack{DiT\\time\(\downarrow\)} &
\shortstack{I2V\\sub.\(\uparrow\)} &
\shortstack{I2V\\back.\(\uparrow\)} &
\shortstack{Total\\score\(\uparrow\)} \\
\midrule
w/o CSG & \ding{55} & 22.1GB & 30s & 0.991 & 0.998 & 6.414 \\
w/o bi-inter. & \ding{51} & 23.2GB & 16s & 0.989 & 0.996 & 6.392 \\
w/o stage-trans & \ding{51} & 20.8GB & 25s & 0.989 & 0.996 & 6.398 \\
\textbf{SwiftI2V} & \ding{51} & \textbf{20.8GB} & \textbf{25s} & \textbf{0.991} & \textbf{0.998} & \textbf{6.424} \\
\bottomrule
\end{tabular}
\end{minipage}
\end{table}

\subsection{Ablation Studies}

\begin{wrapfigure}[14]{r}{0.5\textwidth}
    \centering
    \vspace{-3.5em}
    \includegraphics[width=\linewidth]{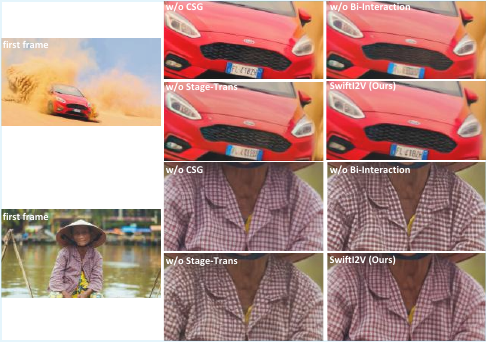}
    \vspace{-1em}
    \caption{\textbf{Ablation studies.} We show qualitative comparisons between different methods.}
    \label{fig:ablation}
\end{wrapfigure}

We ablate SwiftI2V's core components. Unless otherwise specified, all variants are re-trained under their corresponding training protocols with the same optimization budget. Since Stage~I is fixed across ablations, Table~\ref{tab:ablation} reports Stage~II peak memory and DiT runtime, with qualitative results in Figure~\ref{fig:ablation}.
% We perform ablation studies to evaluate how the core components of SwiftI2V affect I2V fidelity and efficiency. Unless otherwise specified, all variants are re-trained under their corresponding training protocols with the same optimization budget. Since the Stage~I input is fixed across ablations, we report Stage~II peak GPU memory (\emph{Peak Mem.}) and diffusion transformer runtime (\emph{DiT Time}) in Table~\ref{tab:ablation} for clarity. Qualitative results are shown in Figure~\ref{fig:ablation}.

\noindent\textbf{Conditional Segment-wise Generation (CSG).}
CSG is designed to keep Stage~II computation and memory within a fixed-size window, enabling streaming inference and improving computational scalability for longer videos. In contrast, \texttt{w/o CSG} removes segmentation and can be viewed as using a full segment window (\ie, \(N{=}0,\, M{=}t{-}1\)), which therefore does \emph{not} support streaming output. Table~\ref{tab:ablation} shows that CSG reduces runtime and peak memory while preserving I2V Subject/Background fidelity and slightly improving total quality. Beyond efficiency and quality, CSG also preserves temporal smoothness across segment boundaries (Appendix~\ref{app:temporal-smoothness}) and improves computational scalability to longer videos (Appendix~\ref{app:long-analysis}).

\noindent\textbf{Bidirectional Interaction.}
We ablate the bidirectional interaction using \texttt{w/o bi-interaction}, where we enforce a causal mask such that the information flow becomes unidirectional, consistent with standard auto-regressive (AR) formulations~\cite{histream,casuvid}. We also enable \textit{key--value (KV) cache} for this AR variant as a standard acceleration strategy. As shown in Table~\ref{tab:ablation}, \texttt{w/o bi-interaction} runs faster but uses more peak memory (due to caching), and it yields a clear drop in both I2V fidelity and overall quality. We attribute this degradation to weaker cross-segment alignment under the causal AR masking, which tends to amplify segmentation-induced discontinuities and error accumulation (Figure~\ref{fig:ablation}; see Appendix~\ref{app:error-accumulation} for a detailed analysis). In contrast, bidirectional interaction enables more adaptive contextual alignment and improves temporal continuity and fidelity.

\noindent\textbf{Stage Transition Training.}
We further ablate the stage transition training (\cf, Section~\ref{sec:stage-trans}) with \texttt{w/o stage-trans}. Removing this strategy consistently degrades quantitative results. Figure~\ref{fig:ablation} shows that Stage~I artifacts persist after refinement, confirming that transition training improves robustness to stage-interface mismatches.

In summary, our ablations validate SwiftI2V's synergistic design: \textbf{CSG} enables bounded-cost streaming inference, \textbf{bidirectional interaction} maintains input-conditioned fidelity, and \textbf{stage transition training} improves two-stage robustness and the quality of generation.

\section{Conclusion}

We presented \textbf{SwiftI2V} to address the efficiency--fidelity dilemma in 2K I2V generation. By decoupling motion modeling from detail synthesis via Conditional Segment-wise Generation (CSG), it avoids the prohibitive costs of end-to-end models. SwiftI2V achieves competitive quality while reducing GPU-time by over \(200\times\) compared to a full-sequence baseline, and lowers the hardware barrier to consumer-grade GPUs. This efficient segment-based paradigm offers a promising direction toward scalable, long-duration, and interactive generative video.

\newpage
\bibliographystyle{plainnat}
\bibliography{swifti2v}

@misc{cinescale,
      title={CineScale: Free Lunch in High-Resolution Cinematic Visual Generation}, 
      author={Haonan Qiu and Ning Yu and Ziqi Huang and Paul Debevec and Ziwei Liu},
      year={2025},
      eprint={2508.15774},
      archivePrefix={arXiv},
      primaryClass={cs.CV},
      url={https://arxiv.org/abs/2508.15774}, 
}

@misc{flashvideo,
      title={FlashVideo: Flowing Fidelity to Detail for Efficient High-Resolution Video Generation}, 
      author={Shilong Zhang and Wenbo Li and Shoufa Chen and Chongjian Ge and Peize Sun and Yida Zhang and Yi Jiang and Zehuan Yuan and Binyue Peng and Ping Luo},
      year={2025},
      eprint={2502.05179},
      archivePrefix={arXiv},
      primaryClass={cs.CV},
      url={https://arxiv.org/abs/2502.05179}, 
}

@misc{diffvsr,
      title={DiffVSR: Revealing an Effective Recipe for Taming Robust Video Super-Resolution Against Complex Degradations}, 
      author={Xiaohui Li and Yihao Liu and Shuo Cao and Ziyan Chen and Shaobin Zhuang and Xiangyu Chen and Yinan He and Yi Wang and Yu Qiao},
      year={2025},
      eprint={2501.10110},
      archivePrefix={arXiv},
      primaryClass={cs.CV},
      url={https://arxiv.org/abs/2501.10110}, 
}

@article{streamdiffvsr,
  title={Stream-DiffVSR: Low-Latency Streamable Video Super-Resolution via Auto-Regressive Diffusion},
  author={Shiu, Hau-Shiang and Lin, Chin-Yang and Wang, Zhixiang and Hsiao, Chi-Wei and Yu, Po-Fan and Chen, Yu-Chih and Liu, Yu-Lun},
  journal={arXiv preprint arXiv:2512.23709},
  year={2025}
}

@article{vdm,
  title={Video diffusion models},
  author={Ho, Jonathan and Salimans, Tim and Gritsenko, Alexey and Chan, William and Norouzi, Mohammad and Fleet, David J},
  journal={Advances in neural information processing systems},
  volume={35},
  pages={8633--8646},
  year={2022}
}

@misc{vldm,
      title={Align your Latents: High-Resolution Video Synthesis with Latent Diffusion Models}, 
      author={Andreas Blattmann and Robin Rombach and Huan Ling and Tim Dockhorn and Seung Wook Kim and Sanja Fidler and Karsten Kreis},
      year={2023},
      eprint={2304.08818},
      archivePrefix={arXiv},
      primaryClass={cs.CV},
      url={https://arxiv.org/abs/2304.08818}, 
}

@inproceedings{ldm,
  title={High-resolution image synthesis with latent diffusion models},
  author={Rombach, Robin and Blattmann, Andreas and Lorenz, Dominik and Esser, Patrick and Ommer, Bj{\"o}rn},
  booktitle={Proceedings of the IEEE/CVF conference on computer vision and pattern recognition},
  pages={10684--10695},
  year={2022}
}

@article{cogvideox,
  title={Cogvideox: Text-to-video diffusion models with an expert transformer},
  author={Yang, Zhuoyi and Teng, Jiayan and Zheng, Wendi and Ding, Ming and Huang, Shiyu and Xu, Jiazheng and Yang, Yuanming and Hong, Wenyi and Zhang, Xiaohan and Feng, Guanyu and others},
  journal={arXiv preprint arXiv:2408.06072},
  year={2024}
}

@article{wan,
      title={Wan: Open and Advanced Large-Scale Video Generative Models}, 
      author={Team Wan and Ang Wang and Baole Ai and Bin Wen and Chaojie Mao and Chen-Wei Xie and Di Chen and Feiwu Yu and Haiming Zhao and Jianxiao Yang and Jianyuan Zeng and Jiayu Wang and Jingfeng Zhang and Jingren Zhou and Jinkai Wang and Jixuan Chen and Kai Zhu and Kang Zhao and Keyu Yan and Lianghua Huang and Mengyang Feng and Ningyi Zhang and Pandeng Li and Pingyu Wu and Ruihang Chu and Ruili Feng and Shiwei Zhang and Siyang Sun and Tao Fang and Tianxing Wang and Tianyi Gui and Tingyu Weng and Tong Shen and Wei Lin and Wei Wang and Wei Wang and Wenmeng Zhou and Wente Wang and Wenting Shen and Wenyuan Yu and Xianzhong Shi and Xiaoming Huang and Xin Xu and Yan Kou and Yangyu Lv and Yifei Li and Yijing Liu and Yiming Wang and Yingya Zhang and Yitong Huang and Yong Li and You Wu and Yu Liu and Yulin Pan and Yun Zheng and Yuntao Hong and Yupeng Shi and Yutong Feng and Zeyinzi Jiang and Zhen Han and Zhi-Fan Wu and Ziyu Liu},
      journal = {arXiv preprint arXiv:2503.20314},
      year={2025}
}

@inproceedings{motioni2v,
  title={Motion-i2v: Consistent and controllable image-to-video generation with explicit motion modeling},
  author={Shi, Xiaoyu and Huang, Zhaoyang and Wang, Fu-Yun and Bian, Weikang and Li, Dasong and Zhang, Yi and Zhang, Manyuan and Cheung, Ka Chun and See, Simon and Qin, Hongwei and others},
  booktitle={ACM SIGGRAPH 2024 Conference Papers},
  pages={1--11},
  year={2024}
}

@article{i2vgen,
  title={I2vgen-xl: High-quality image-to-video synthesis via cascaded diffusion models},
  author={Zhang, Shiwei and Wang, Jiayu and Zhang, Yingya and Zhao, Kang and Yuan, Hangjie and Qin, Zhiwu and Wang, Xiang and Zhao, Deli and Zhou, Jingren},
  journal={arXiv preprint arXiv:2311.04145},
  year={2023}
}

@inproceedings{guo2024make,
  title={Make a cheap scaling: A self-cascade diffusion model for higher-resolution adaptation},
  author={Guo, Lanqing and He, Yingqing and Chen, Haoxin and Xia, Menghan and Cun, Xiaodong and Wang, Yufei and Huang, Siyu and Zhang, Yong and Wang, Xintao and Chen, Qifeng and others},
  booktitle={European conference on computer vision},
  pages={39--55},
  year={2024},
  organization={Springer}
}

@inproceedings{turbo2k,
  title={Turbo2k: Towards ultra-efficient and high-quality 2k video synthesis},
  author={Ren, Jingjing and Li, Wenbo and Wang, Zhongdao and Sun, Haoze and Liu, Bangzhen and Chen, Haoyu and Xu, Jiaqi and Li, Aoxue and Zhang, Shifeng and Shao, Bin and others},
  booktitle={Proceedings of the IEEE/CVF International Conference on Computer Vision},
  pages={18155--18165},
  year={2025}
}

@article{wang2023videolcm,
  title={Videolcm: Video latent consistency model},
  author={Wang, Xiang and Zhang, Shiwei and Zhang, Han and Liu, Yu and Zhang, Yingya and Gao, Changxin and Sang, Nong},
  journal={arXiv preprint arXiv:2312.09109},
  year={2023}
}

@inproceedings{upscaleavideo,
  title={Upscale-a-video: Temporal-consistent diffusion model for real-world video super-resolution},
  author={Zhou, Shangchen and Yang, Peiqing and Wang, Jianyi and Luo, Yihang and Loy, Chen Change},
  booktitle={Proceedings of the IEEE/CVF Conference on Computer Vision and Pattern Recognition},
  pages={2535--2545},
  year={2024}
}

@inproceedings{casuvid,
  title={From slow bidirectional to fast autoregressive video diffusion models},
  author={Yin, Tianwei and Zhang, Qiang and Zhang, Richard and Freeman, William T and Durand, Fredo and Shechtman, Eli and Huang, Xun},
  booktitle={Proceedings of the Computer Vision and Pattern Recognition Conference},
  pages={22963--22974},
  year={2025}
}

@article{histream,
  title={HiStream: Efficient High-Resolution Video Generation via Redundancy-Eliminated Streaming},
  author={Qiu, Haonan and Liu, Shikun and Zhou, Zijian and An, Zhaochong and Ren, Weiming and Liu, Zhiheng and Schult, Jonas and He, Sen and Chen, Shoufa and Cong, Yuren and others},
  journal={arXiv preprint arXiv:2512.21338},
  year={2025}
}

@misc{ltx2,
      title={LTX-2: Efficient Joint Audio-Visual Foundation Model}, 
      author={Yoav HaCohen and Benny Brazowski and Nisan Chiprut and Yaki Bitterman and Andrew Kvochko and Avishai Berkowitz and Daniel Shalem and Daphna Lifschitz and Dudu Moshe and Eitan Porat and Eitan Richardson and Guy Shiran and Itay Chachy and Jonathan Chetboun and Michael Finkelson and Michael Kupchick and Nir Zabari and Nitzan Guetta and Noa Kotler and Ofir Bibi and Ori Gordon and Poriya Panet and Roi Benita and Shahar Armon and Victor Kulikov and Yaron Inger and Yonatan Shiftan and Zeev Melumian and Zeev Farbman},
      year={2026},
      eprint={2601.03233},
      archivePrefix={arXiv},
      primaryClass={cs.CV},
      url={https://arxiv.org/abs/2601.03233}, 
}

@misc{dit,
      title={Scalable Diffusion Models with Transformers}, 
      author={William Peebles and Saining Xie},
      year={2023},
      eprint={2212.09748},
      archivePrefix={arXiv},
      primaryClass={cs.CV},
      url={https://arxiv.org/abs/2212.09748}, 
}

@misc{lightx2v,
     author = {LightX2V},
     title = {LightX2V: Light Video Generation Inference Framework},
     year = {2025},
     publisher = {GitHub},
     journal = {GitHub repository},
     howpublished = {\url{https://github.com/ModelTC/lightx2v}},
}

@article{teacherfocing,
  title={A learning algorithm for continually running fully recurrent neural networks},
  author={Williams, Ronald J and Zipser, David},
  journal={Neural computation},
  volume={1},
  number={2},
  pages={270--280},
  year={1989},
  publisher={MIT Press One Rogers Street, Cambridge, MA 02142-1209, USA journals-info~…}
}

@article{openvid,
  title={OpenVid-1M: A Large-Scale High-Quality Dataset for Text-to-video Generation},
  author={Nan, Kepan and Xie, Rui and Zhou, Penghao and Fan, Tiehan and Yang, Zhenheng and Chen, Zhijie and Li, Xiang and Yang, Jian and Tai, Ying},
  journal={arXiv preprint arXiv:2407.02371},
  year={2024}
}

@article{ultravideo,
  title={UltraVideo: High-Quality UHD Video Dataset with Comprehensive Captions},
  author={Xue, Zhucun and Zhang, Jiangning and Hu, Teng and He, Haoyang and Chen, Yinan and Cai, Yuxuan and Wang, Yabiao and Wang, Chengjie and Liu, Yong and Li, Xiangtai and Tao, Dacheng}, 
  journal={arXiv preprint arXiv:2506.13691},
  year={2025}
}

@article{vbench++,
    title={{VBench++}: Comprehensive and Versatile Benchmark Suite for Video Generative Models},
    author={Huang, Ziqi and Zhang, Fan and Xu, Xiaojie and He, Yinan and Yu, Jiashuo and Dong, Ziyue and Ma, Qianli and Chanpaisit, Nattapol and Si, Chenyang and Jiang, Yuming and Wang, Yaohui and Chen, Xinyuan and Chen, Ying-Cong and Wang, Limin and Lin, Dahua and Qiao, Yu and Liu, Ziwei},
    journal={IEEE Transactions on Pattern Analysis and Machine Intelligence}, 
    year={2025},
    doi={10.1109/TPAMI.2025.3633890}
}

@misc{turbovsr,
      title={TurboVSR: Fantastic Video Upscalers and Where to Find Them}, 
      author={Zhongdao Wang and Guodongfang Zhao and Jingjing Ren and Bailan Feng and Shifeng Zhang and Wenbo Li},
      year={2025},
      eprint={2506.23618},
      archivePrefix={arXiv},
      primaryClass={cs.CV},
      url={https://arxiv.org/abs/2506.23618}, 
}

@software{diffsynth_studio,
  title={DiffSynth-Studio: An Open-Source Diffusion Model Engine},
  author={ModelScope},
  year={2024},
  url={https://github.com/modelscope/DiffSynth-Studio},
  note={GitHub repository}
}

@misc{hunyuan1.5,
      title={HunyuanVideo 1.5 Technical Report}, 
      author={Tencent Hunyuan},
      year={2025},
      eprint={2511.18870},
      archivePrefix={arXiv},
      primaryClass={cs.CV},
      url={https://arxiv.org/abs/2511.18870}, 
}

@misc{seedvr,
      title={SeedVR2: One-Step Video Restoration via Diffusion Adversarial Post-Training}, 
      author={Jianyi Wang and Shanchuan Lin and Zhijie Lin and Yuxi Ren and Meng Wei and Zongsheng Yue and Shangchen Zhou and Hao Chen and Yang Zhao and Ceyuan Yang and Xuefeng Xiao and Chen Change Loy and Lu Jiang},
      year={2025},
      eprint={2506.05301},
      archivePrefix={arXiv},
      primaryClass={cs.CV},
      url={https://arxiv.org/abs/2506.05301}, 
}

@misc{star,
      title={STAR: Spatial-Temporal Augmentation with Text-to-Video Models for Real-World Video Super-Resolution}, 
      author={Rui Xie and Yinhong Liu and Penghao Zhou and Chen Zhao and Jun Zhou and Kai Zhang and Zhenyu Zhang and Jian Yang and Zhenheng Yang and Ying Tai},
      year={2025},
      eprint={2501.02976},
      archivePrefix={arXiv},
      primaryClass={cs.CV},
      url={https://arxiv.org/abs/2501.02976}, 
}

@article{lora,
  title={Lora: Low-rank adaptation of large language models.},
  author={Hu, Edward J and Shen, Yelong and Wallis, Phillip and Allen-Zhu, Zeyuan and Li, Yuanzhi and Wang, Shean and Wang, Lu and Chen, Weizhu and others},
  journal={ICLR},
  volume={1},
  number={2},
  pages={3},
  year={2022}
}

@article{adamw,
  title={Decoupled weight decay regularization},
  author={Loshchilov, Ilya and Hutter, Frank},
  journal={arXiv preprint arXiv:1711.05101},
  year={2017}
}

@misc{magi-1,
      title={MAGI-1: Autoregressive Video Generation at Scale}, 
      author={Sand. ai and Hansi Teng and Hongyu Jia and Lei Sun and Lingzhi Li and Maolin Li and Mingqiu Tang and Shuai Han and Tianning Zhang and W. Q. Zhang and Weifeng Luo and Xiaoyang Kang and Yuchen Sun and Yue Cao and Yunpeng Huang and Yutong Lin and Yuxin Fang and Zewei Tao and Zheng Zhang and Zhongshu Wang and Zixun Liu and Dai Shi and Guoli Su and Hanwen Sun and Hong Pan and Jie Wang and Jiexin Sheng and Min Cui and Min Hu and Ming Yan and Shucheng Yin and Siran Zhang and Tingting Liu and Xianping Yin and Xiaoyu Yang and Xin Song and Xuan Hu and Yankai Zhang and Yuqiao Li},
      year={2025},
      eprint={2505.13211},
      archivePrefix={arXiv},
      primaryClass={cs.CV},
      url={https://arxiv.org/abs/2505.13211}, 
}

@article{nova,
  title={Autoregressive Video Generation without Vector Quantization},
  author={Deng, Haoge and Pan, Ting and Diao, Haiwen and Luo, Zhengxiong and Cui, Yufeng and Lu, Huchuan and Shan, Shiguang and Qi, Yonggang and Wang, Xinlong},
  journal={arXiv preprint arXiv:2412.14169},
  year={2024}
}

\clearpage
\appendix

% ---- Manual Appendix ToC (robust under NeurIPS template) ----
\section*{Appendix}
\setlength{\tabcolsep}{0pt}
\renewcommand{\arraystretch}{1.25}
\noindent\begin{tabularx}{\linewidth}{@{}l X r@{}}
\textbf{\ref{app:limitations}} & \textbf{Limitations and Broader Impacts} \dotfill & \pageref{app:limitations} \\
\textbf{\ref{app:impl}} & \textbf{More Implementation Details} \dotfill & \pageref{app:impl} \\
\quad \ref{app:impl-train} & Training Details \dotfill & \pageref{app:impl-train} \\
\quad \ref{app:impl-infer} & Inference Details \dotfill & \pageref{app:impl-infer} \\
\textbf{\ref{app:exp}} & \textbf{Additional Experimental Results} \dotfill & \pageref{app:exp} \\
\multicolumn{3}{@{}l}{\quad\textit{--- Benchmark Results ---}} \\
\quad \ref{app:full-results} & Full VBench-I2V Results \dotfill & \pageref{app:full-results} \\
\quad \ref{app:cinescale-subset} & Subset Evaluation Results \dotfill & \pageref{app:cinescale-subset} \\
\multicolumn{3}{@{}l}{\quad\textit{--- Extended Settings and Applications ---}} \\
\quad \ref{app:long-analysis} & Scalability Analysis \dotfill & \pageref{app:long-analysis} \\
\quad \ref{app:4090} & Experiments on Consumer GPUs \dotfill & \pageref{app:4090} \\
\quad \ref{app:streaming} & Streaming Generation with CSG \dotfill & \pageref{app:streaming} \\
\multicolumn{3}{@{}l}{\quad\textit{--- In-depth Analysis ---}} \\
\quad \ref{app:mn-selection} & Ablation on Segmentation Hyperparameters $M$ and $N$ \dotfill & \pageref{app:mn-selection} \\
\quad \ref{app:stage-trans} & Analysis of Stage Transition \dotfill & \pageref{app:stage-trans} \\
\quad \ref{app:vae-fidelity} & VAE Reconstruction Fidelity \dotfill & \pageref{app:vae-fidelity} \\
\quad \ref{app:temporal-smoothness} & Temporal Smoothness across Segment Boundaries \dotfill & \pageref{app:temporal-smoothness} \\
\quad \ref{app:error-accumulation} & Analysis of Cross-Segment Error Accumulation \dotfill & \pageref{app:error-accumulation} \\
\quad \ref{app:dynamic-degree} & Analysis of Dynamic Degree \dotfill & \pageref{app:dynamic-degree} \\
\textbf{\ref{app:visual}} & \textbf{More 2K I2V Visual Results of SwiftI2V} \dotfill & \pageref{app:visual} \\
\end{tabularx}

\vspace{3em}

\section{Limitations and Broader Impacts}
\label{app:limitations}

\noindent\textbf{Limitations.}
SwiftI2V substantially improves the efficiency and deployability of 2K I2V generation, and its Conditional Segment-wise Generation further supports streaming generation. However, it still does not achieve strict real-time synthesis. Reaching real-time 2K video generation remains an important direction that may require further advances in model compression, caching, scheduling, and hardware-aware inference.

Another practical consideration is system integration. SwiftI2V deliberately decouples low-resolution motion generation and high-resolution detail synthesis, allowing each stage to use a backbone and resolution setting suited to its role. This modular design improves scalability and fidelity, but deploying two specialized stages can be more involved than serving a single monolithic model, especially when optimizing memory sharing, batching, and model loading in production systems. We view this as an engineering trade-off for efficient high-resolution I2V, and future work can further simplify deployment through unified distillation, shared components, or more integrated inference runtimes.

\noindent\textbf{Broader Impacts.}
By reducing the compute and memory cost of high-resolution I2V synthesis, SwiftI2V may broaden access to video creation and research tools, enable faster prototyping and visualization workflows, and reduce the energy cost per generated sample compared with less efficient high-resolution pipelines. These benefits are most relevant for responsible applications such as content creation, education, design, and research.

At the same time, more efficient I2V generation may also increase risks associated with synthetic media, including deceptive or non-consensual content, impersonation, misinformation, or biased generations. Because the method is conditioned on an input image, responsible deployment should include consent-aware use policies for identifiable people, clear disclosure of synthetic content, provenance or watermarking mechanisms when applicable, and abuse monitoring in downstream systems.

\section{More Implementation Details}
\label{app:impl}

\subsection{Training Details}
\label{app:impl-train}

All experiments are implemented with the DiffSynth-Studio~\cite{diffsynth_studio} framework. We use bicubic interpolation for spatial resizing in data pre-processing and for constructing low-resolution videos.
For both stages, we optimize \textit{only} the DiT; other components in the pipeline (\eg, VAE and text encoder) are kept frozen.

\noindent\textbf{Stage I:}
Concretely, 360P-LoRA~\cite{lora} is inserted into the DiT attention projections and MLP layers with rank \(r=128\).
We optimize with AdamW~\cite{adamw} using learning rate \(\mathrm{lr}=1\times 10^{-4}\) and a constant learning-rate schedule.

\noindent\textbf{Stage II:}
We train on fixed \(2560\times 1408\) videos with 81 frames. To incorporate the low-resolution reference via channel-wise concatenation, we expand the DiT patch-embedding input channels from 48 to 96. The expanded patch embedding is initialized by copying pretrained weights for the first 48 channels and zero-initializing the additional 48 channels, which stabilizes training at the stage-transition interface.
We fully fine-tune the DiT with AdamW using learning rate \(\mathrm{lr}=5\times 10^{-6}\) and a constant learning-rate schedule.
For robustness to different segmentations, we randomize the segment hyperparameters by sampling \((M,N)\) uniformly from the four combinations \(M\in\{2,3\}\) and \(N\in\{1,2\}\) at each iteration.
Following Section~\ref{sec:stage-trans}, we train Stage~II using pre-generated stage-transition samples, and mix them with downsampled counterparts at a ratio of \(7:3\) (generated:downsampled).

\subsection{Inference Details}
\label{app:impl-infer}

\noindent\textbf{Stage I:}
During inference, we load two LoRA adapters into the DiT of Wan2.1-I2V-14B-480P~\cite{wan} with \(\alpha=1\): a few-step distillation LoRA~\cite{lightx2v} and our 360P-LoRA.
We sample at 360P (\(640\times 352\)) resolution with 4 denoising steps and without CFG.

\noindent\textbf{Stage II:}
For Stage~II inference, we apply CSG with fixed \((M,N)=(3,1)\).
After VAE encoding, the latent sequence contains 21 temporal blocks; the first block corresponds to the input-image anchor and is kept fixed, while the remaining 20 blocks are denoised.
We partition these 20 blocks into 7 sequential segments (6 segments of length 3 and a shorter final segment).
For each segment, we run 4 denoising steps without CFG. To further accelerate inference and reduce peak memory, Stage~II employs tiled VAE encoding/decoding.

\vspace{3em}
\section{Additional Experimental Results}
\label{app:exp}

\subsection{Full VBench-I2V results}
\label{app:full-results}

\begin{figure}[H]
    \centering
    \includegraphics[width=0.6\linewidth]{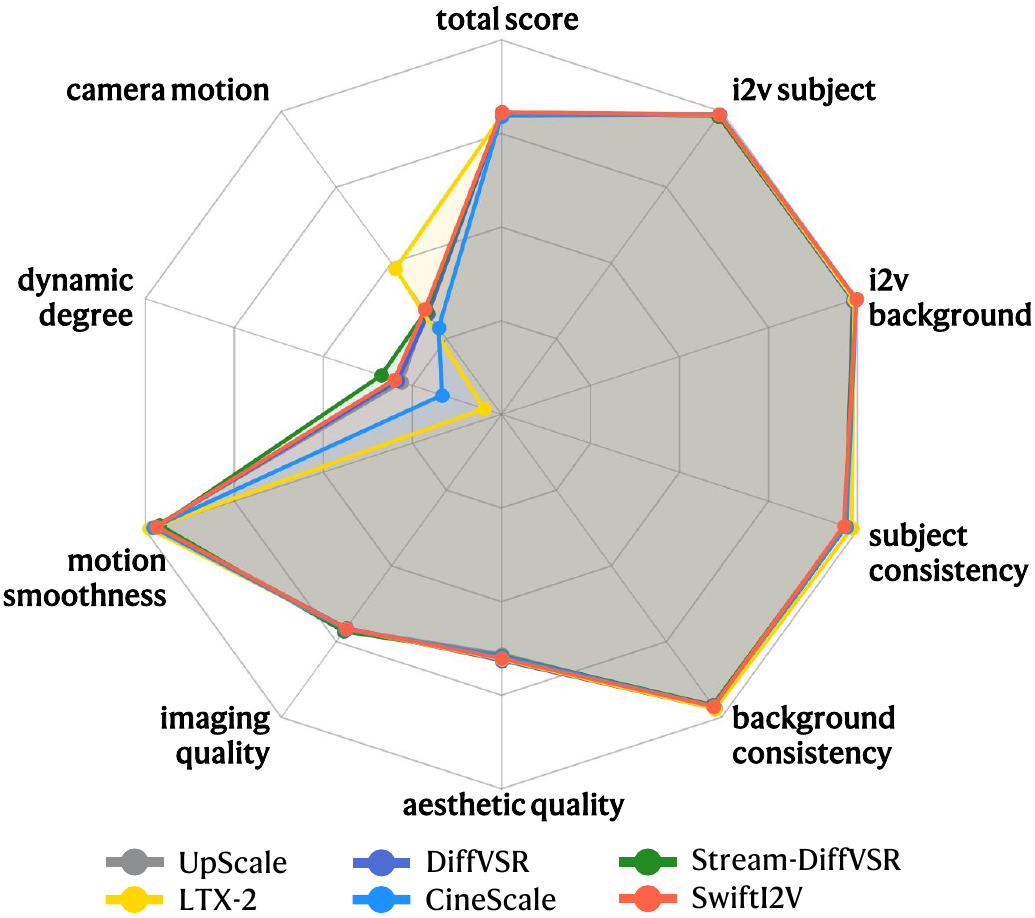}
    \caption{\textbf{VBench-I2V score visualization.} We show all dimension scores here.}
    \label{fig:all_result}
\end{figure}

Figure~\ref{fig:all_result} reports the complete VBench-I2V evaluation results of SwiftI2V and other baselines (DiffVSR~\cite{diffvsr}, Stream-DiffVSR~\cite{streamdiffvsr}, LTX-2~\cite{ltx2}, CineScale~\cite{cinescale}) across both the total score and individual dimensions. SwiftI2V achieves the best total score among the compared methods. It performs consistently well on I2V-related criteria and attains competitive performance on aesthetic-related metrics. Meanwhile, SwiftI2V also demonstrates a high dynamic degree.

\subsection{Subset Evaluation Results}
\label{app:cinescale-subset}

In the main comparison in Section~\ref{sec:main-comparison}, we include CineScale~\cite{cinescale} as a representative end-to-end 2K I2V baseline. However, under the 2K resolution and 81-frame setting, CineScale incurs substantially higher inference cost than other methods: even with 8 H800 GPUs in parallel, generating a single sample takes approximately 50 minutes. Due to limited computational resources, we cannot afford to evaluate CineScale on the full official VBench-I2V test set (1118 samples). Therefore, we construct a randomly sampled subset for a feasible yet informative comparison.

Specifically, we randomly select a subset of 63 samples from the full test set. \textit{While ensuring randomness, we also keep the ratio of samples associated with each evaluation dimension close to that of the original test set}, so as to reduce potential sampling bias. Except for using the subset instead of the full set, all evaluation protocols remain identical to those in the main paper.

Table~\ref{tab:subset} reports the VBench-I2V results of SwiftI2V and CineScale on this subset. SwiftI2V consistently maintains a high Dynamic Degree while outperforming CineScale on the overall score and all I2V-related metrics on this subset. These results further support the conclusions in the main paper and indicate that, compared with CineScale, SwiftI2V can generate 2K I2V videos more efficiently while achieving both strong motion dynamics and high input fidelity.

\begin{table}[H]
\centering
\small
\caption{Comparison of the results of CineScale and SwiftI2V on the VBench-I2V subset.}
\label{tab:subset}
\begin{tabularx}{0.7\linewidth}{lCCCC}
\toprule
\multirow{2}{*}{Model} & Total & I2V & I2V & Dynamic \\
 & Score\(\uparrow\) & Subject\(\uparrow\) & Background\(\uparrow\) & Degree\(\uparrow\) \\
\midrule
CineScale & 6.3638 & 0.9924 & 0.9973 & 0.1667 \\
\textbf{SwiftI2V} & \textbf{6.4284} & \textbf{0.9927} & \textbf{0.9983} & \textbf{0.2917} \\
\bottomrule
\end{tabularx}
\end{table}

To further validate the reliability of the performance improvements on this 63-sample subset, we conduct a sample-wise paired statistical significance test between SwiftI2V and CineScale. We use the Wilcoxon signed-rank test for continuous metrics (I2V Subject and I2V Background) and McNemar's exact test for the binary metric (Dynamic Degree). The results are summarized in Table~\ref{tab:subset_significance}.

\begin{table}[H]
\centering
\small
\caption{Statistical significance test between SwiftI2V and CineScale on the VBench-I2V subset.}
\label{tab:subset_significance}
\begin{tabularx}{0.8\linewidth}{lCCC}
\toprule
Metric & SwiftI2V & CineScale & $p$-value \\
\midrule
I2V Subject $\uparrow$ & $0.993 \pm 0.005$ & $0.992 \pm 0.005$ & $0.944$ \\
I2V Background $\uparrow$ & $\mathbf{0.998 \pm 0.001}$ & $0.997 \pm 0.002$ & $0.042^*$ \\
Dynamic Degree $\uparrow$ & $\mathbf{0.292 \pm 0.455}$ & $0.167 \pm 0.373$ & $0.375$ ($g=0.30$) \\
\bottomrule
\end{tabularx}
\end{table}

The results indicate that SwiftI2V achieves comparable image fidelity to CineScale (I2V Subject, $p=0.944$) while demonstrating a statistically significant advantage in background consistency (I2V Background, $p < 0.05$). Regarding the Dynamic Degree, SwiftI2V increases the proportion of dynamic videos from $16.7\%$ to $29.2\%$. Although McNemar's test does not reach statistical significance ($p=0.375$) due to the limited number of discordant pairs, the effect size reaches Cohen's $g=0.30$ (a large effect), indicating a practically meaningful improvement. This trend is also evident in our qualitative comparisons, where CineScale occasionally fails to generate any meaningful motion. Overall, these subset results support that SwiftI2V achieves a $202\times$ acceleration without sacrificing image fidelity, while showing advantages in background consistency and motion dynamics.

\begin{figure}[H]
  \centering
  \includegraphics[width=0.6\linewidth]{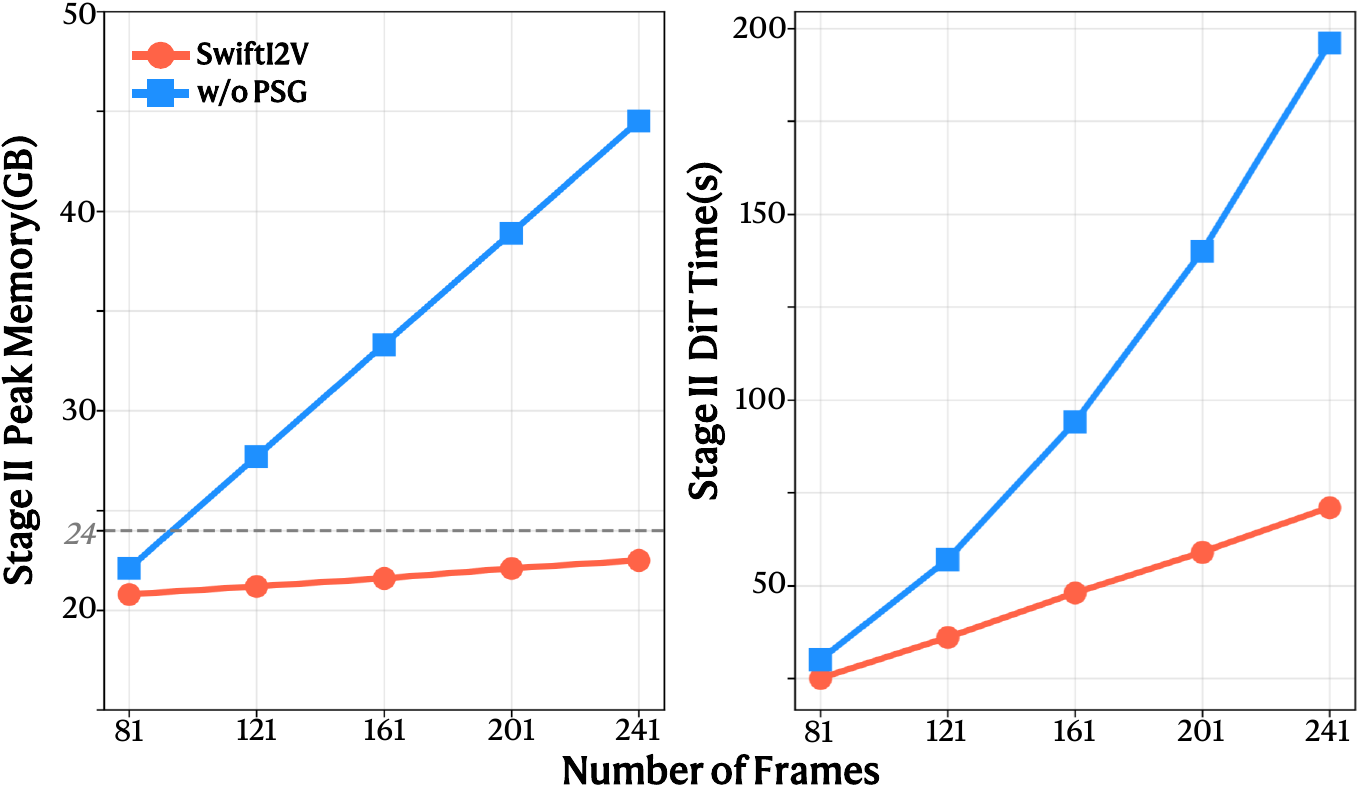}
  \caption{\textbf{Scaling behavior at 2K resolution.} Left: Stage~II peak GPU memory vs.\ number of frames. Right: Stage~II DiT generating time vs.\ number of frames. SwiftI2V keeps peak memory below 24\,GB even at 241 frames and exhibits near-linear time growth, while removing CSG leads to rapidly increasing memory and runtime.}
  \label{fig:scaling_csg}
\end{figure}

\subsection{Scalability Analysis}
\label{app:long-analysis}

This subsection evaluates different methods' scalability with respect to the target video length, with emphasis on how Conditional Segment-wise Generation (CSG) affects GPU memory and runtime. We fix the output resolution to \(2560\times 1408\) and vary the generated length \(T\in\{81,121,161,201,241\}\) frames. All measurements are conducted on a single NVIDIA H800 GPU. We report the peak GPU memory footprint of Stage~II and the DiT time of Stage~II for clear comparison. We compare \textbf{SwiftI2V} against the variant that removes CSG (\texttt{w/o CSG}). The results are summarized in Figure~\ref{fig:scaling_csg}.

As shown in Figure~\ref{fig:scaling_csg}, CSG yields a highly controllable memory footprint: even at \(T=241\) frames, the Stage~II peak memory remains below \(24\,\mathrm{GB}\). Moreover, the Stage~II sampling time increases approximately linearly with \(T\). This trend is consistent with CSG's design: under a fixed spatial resolution, CSG enforces a constant token budget per segment, so the overall computation is dominated by the number of segments \(S\), leading to near-linear scaling in \(S\) (and thus in \(T\)).

In contrast, removing CSG requires full-temporal denoising over the entire sequence, for which the token count---and consequently the attention footprint and compute---grows with \(T\). As a result, both the peak memory and runtime increase substantially faster as the target length increases, reflecting the unfavorable scaling of full-temporal denoising at 2K resolution.

Overall, these results highlight CSG as a key enabler for computationally practical long-video generation at high resolution: it provides a stable memory bound and near-linear time scaling with the number of segments \(S\), making Stage~II inference feasible and predictable for substantially longer outputs.

\subsection{Experiments on Consumer GPUs}
\label{app:4090}

To validate the practicality of SwiftI2V beyond datacenter GPUs, we further deploy our full pipeline on a single consumer-grade GPU, \textbf{NVIDIA RTX 4090 (24GB)}. All experiments are conducted under the DiffSynth-Studio~\cite{diffsynth_studio} framework with the same default 2K I2V setting as in the main paper (\ie, \textbf{2560\(\times\)1408} resolution and \textbf{81} frames).

For Stage~I, we enable \texttt{cpu\_offload} in DiffSynth-Studio to satisfy the memory constraint on a 24GB GPU, which allows the large motion backbone to run with acceptable peak VRAM usage. Thanks to our Conditional Segment-wise Generation (CSG), Stage~II is \emph{highly memory-controllable} with a bounded token budget. As a result, Stage~II can be deployed \textbf{directly on RTX 4090}, while still maintaining stable 2K generation quality.

\begin{figure}[H]
  \centering
  \includegraphics[width=\linewidth]{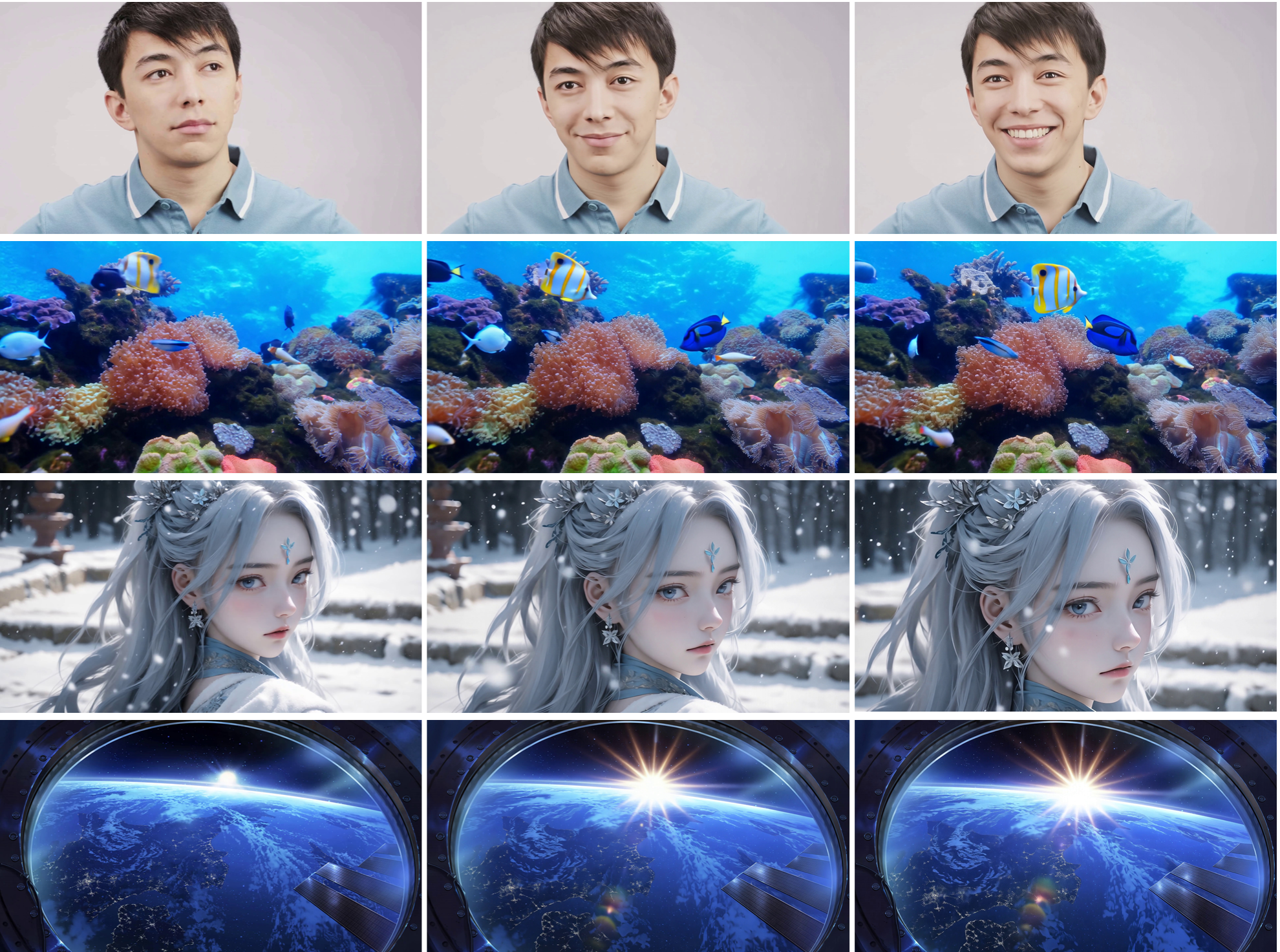}
  \caption{Representative 2K I2V samples generated by SwiftI2V \textbf{on a single NVIDIA RTX 4090 (24GB)}.}
  \label{fig:4090}
\end{figure}

With the above settings, SwiftI2V completes a full high-quality 2K I2V sample in about \textbf{380s} on a single RTX 4090. (\textit{Note: due to hardware/software constraints in our environment, we cannot enable \texttt{flash\_attn} as an acceleration mechanism on RTX 4090. We expect the runtime can be further reduced with \texttt{flash\_attn} or other optimized attention kernels}.) We also provide representative 2K I2V examples generated on RTX~4090 in Figure~\ref{fig:4090}.

Overall, SwiftI2V substantially lowers the hardware barrier of practical high-resolution I2V generation and enables accessible 2K deployment in commodity settings.

\subsection{Streaming Generation with CSG}
\label{app:streaming}

This subsection supplements Section~\ref{sec:csg} with a deployment-oriented study of \emph{streaming output} in Stage~II. In this study, we assume the Stage~I motion reference is available and focus on the Stage~II \emph{generation-to-output} path (DiT denoising of 2K latents \(\rightarrow\) VAE decoding \(\rightarrow\) video saving). We report \emph{time-to-first-viewable-output} and the \emph{incremental delivery rate}. These measurements provide an actionable baseline for building future low-latency 2K I2V systems on top of SwiftI2V.

In the standard sequential execution, Stage~II completes DiT denoising before running VAE decoding and saving. Under our default setting, DiT denoising takes \(25\,\mathrm{s}\) and VAE decoding takes \(22\,\mathrm{s}\). Including video materialization overheads, the end-to-end wall-clock time is \(\sim 50\,\mathrm{s}\), and viewable frames are produced only near the end.

To enable streaming output, we implement a simple pipeline that overlaps DiT denoising and VAE decoding at CSG segment boundaries. Since concurrent DiT+VAE execution on one GPU is impractical due to compute contention, we deploy DiT and VAE on two NVIDIA H800 GPUs. Once DiT finishes denoising a CSG segment, we transfer the corresponding latent segment to the decoder GPU and put it in a queue for decoding. Decoding and output are performed in blocks of 4 frames, each written out as soon as it becomes available.

As shown in Table~\ref{tab:streaming_csg}, DiT produces the first segment latent after \(7.4\,\mathrm{s}\), and \textbf{the first decoded 4-frame block becomes available after \(\mathbf{8.5}\,\mathbf{s}\)} (time-to-first-viewable-output). The full video finishes decoding and output at \(\sim 30\,\mathrm{s}\), close to the ideal pipelining behavior where total time approaches ``DiT latency + full VAE decoding.'' \textbf{During streaming, the system outputs one 4-frame block every \(\mathbf{1.2}\,\mathbf{s}\) on average, \ie, an incremental delivery rate of \(\approx \mathbf{3.33}\) fps.}

Overall, these measurements substantiate the claim in Section~\ref{sec:csg} that CSG not only bounds the per-step 2K token budget for scalable refinement, but also enables low-latency, incremental delivery by making segment-wise decoding possible. While the demonstrated throughput is below real-time playback and currently requires a two-GPU deployment, the results provide a concrete baseline and suggest a promising direction for future low-latency 2K I2V systems (\eg, via faster decoding, better overlap, or dedicated accelerators).

\begin{table}[t]
\centering
\caption{\textbf{Stage~II streaming output enabled by CSG (Stage-II-only timing).}
``First latent'': time until DiT finishes the first CSG segment.
``First output'': time-to-first-viewable-output (first decoded 4-frame block).
``Output FPS'': average incremental delivery rate during streaming.}
\small
\setlength{\tabcolsep}{3.5pt}
\begin{tabular}{lcccc}
\toprule
Setting & First latent & First output & Full output & Output FPS \\
\midrule
Sequential & -- & \(\sim 50\,\mathrm{s}\) & \(\sim 50\,\mathrm{s}\) & -- \\
Streaming  & \(7.4\,\mathrm{s}\) & \(8.5\,\mathrm{s}\) & \(\sim 30\,\mathrm{s}\) & \(3.33\) \\
\bottomrule
\end{tabular}
\label{tab:streaming_csg}
\end{table}

\subsection{Ablation on Segmentation Hyperparameters \texorpdfstring{$M$}{M} and \texorpdfstring{$N$}{N}}
\label{app:mn-selection}

The segmentation hyperparameters \(M\) (number of noisy blocks denoised per step) and \(N\) (number of conditioning blocks that anchor the segment) directly determine the per-step token budget and the amount of cross-segment context available to CSG. To understand their impact, we sweep \(M\in\{2,3\}\) and \(N\in\{1,2\}\) on the VBench-I2V subset while keeping all other settings fixed, and report the results in Table~\ref{tab:mn-ablation}.

\begin{table}[h]
\centering
\small
\setlength{\tabcolsep}{6pt}
\caption{Ablation on the segmentation hyperparameters \(M\) and \(N\) of CSG. ``Time'' denotes the per-sample inference time of Stage~II at \(2560\times 1408\).}
\label{tab:mn-ablation}
\begin{tabular}{cc|c|cccccc}
\toprule
\(M\) & \(N\) & Time & I2V Subj. & I2V Bg. & Subj. Cons. & Motion Smooth. & Dyn. Deg. & Total \\
\midrule
\textbf{3} & \textbf{1} & \textbf{25s} & \textbf{0.9910} & \textbf{0.9975} & \textbf{0.9675} & \textbf{0.9885} & \textbf{0.3009} & \textbf{6.4243} \\
2 & 1 & 26s & 0.9908 & 0.9975 & 0.9660 & 0.9871 & 0.2967 & 6.4107 \\
3 & 2 & 33s & 0.9909 & 0.9976 & 0.9665 & 0.9876 & 0.2967 & 6.4158 \\
2 & 2 & 37s & 0.9908 & 0.9975 & 0.9633 & 0.9854 & 0.2886 & 6.3916 \\
\bottomrule
\end{tabular}
\end{table}

We make the following observations. \textbf{(i)} Increasing the number of noisy blocks per step from \(M{=}2\) to \(M{=}3\) is consistently beneficial at a fixed \(N\): with \(N{=}1\), the total score improves from 6.4107 to 6.4243 while per-sample time even decreases slightly (26s vs.\ 25s) because fewer segmentation steps are needed to cover the full video. A similar trend holds for \(N{=}2\). This indicates that within the token budget allowed by a 2K denoising step, a moderately larger \(M\) amortizes the segmentation overhead more efficiently and yields slightly better cross-segment coherence. \textbf{(ii)} Enlarging the conditioning window from \(N{=}1\) to \(N{=}2\) does not translate into overall quality gains; although it slightly improves I2V Background, it degrades subject consistency (0.9675\(\to\)0.9665 with \(M{=}3\)), motion smoothness (0.9885\(\to\)0.9876), Dynamic Degree (0.3009\(\to\)0.2967), and total score while incurring higher latency. We attribute this to the image-anchored bidirectional interaction in CSG already providing sufficient global context through \(\mathbf{z}_{\mathrm{ref}}\) and the first conditioning block; doubling the conditioning blocks enlarges the attention span but mainly dilutes the anchoring effect, slightly suppressing motion magnitude without adding useful information. \textbf{(iii)} \(N{=}2\) also incurs a non-trivial efficiency cost (e.g., 25s\(\to\)33s at \(M{=}3\) and 26s\(\to\)37s at \(M{=}2\)) since the per-step token budget grows accordingly.

Overall, \((M, N){=}(3, 1)\) provides the best overall quality--efficiency trade-off, and we therefore adopt it as the default configuration throughout our experiments.

\subsection{Analysis of Stage Transition}
\label{app:stage-trans}

This section further analyzes the stage transition strategy in Section~\ref{sec:stage-trans}.
Since Stage~I and Stage~II are trained separately, Stage~II inevitably faces a train--test mismatch:
during inference, its input comes from Stage~I and may contain typical artifacts (\eg, VAE distortions or low-resolution temporal flicker),
whereas during training it is fed with clean low-resolution videos obtained by direct downsampling.
Such a mismatch causes Stage~II to amplify artifacts in the high-resolution outputs.
Our objective is to inject Stage~I-like artifacts into the low-resolution training inputs \emph{\textbf{without} changing the motion correspondence} between \(V^{LR}\) and \(V^{HR}\),
so as to preserve the motion--detail decoupled design.

Given \(V^{HR}\), we downsample it to \(V^{LR}\) (360P), add noise with strength \(\sigma\in[0,1]\),
and denoise it with Stage~I for \(S\) steps to get \(\tilde V^{LR}\).
\(\sigma=0\) means no noise and \(\sigma=1\) corresponds to starting from pure noise.

\begin{figure}[t]
  \centering
  \includegraphics[width=\linewidth]{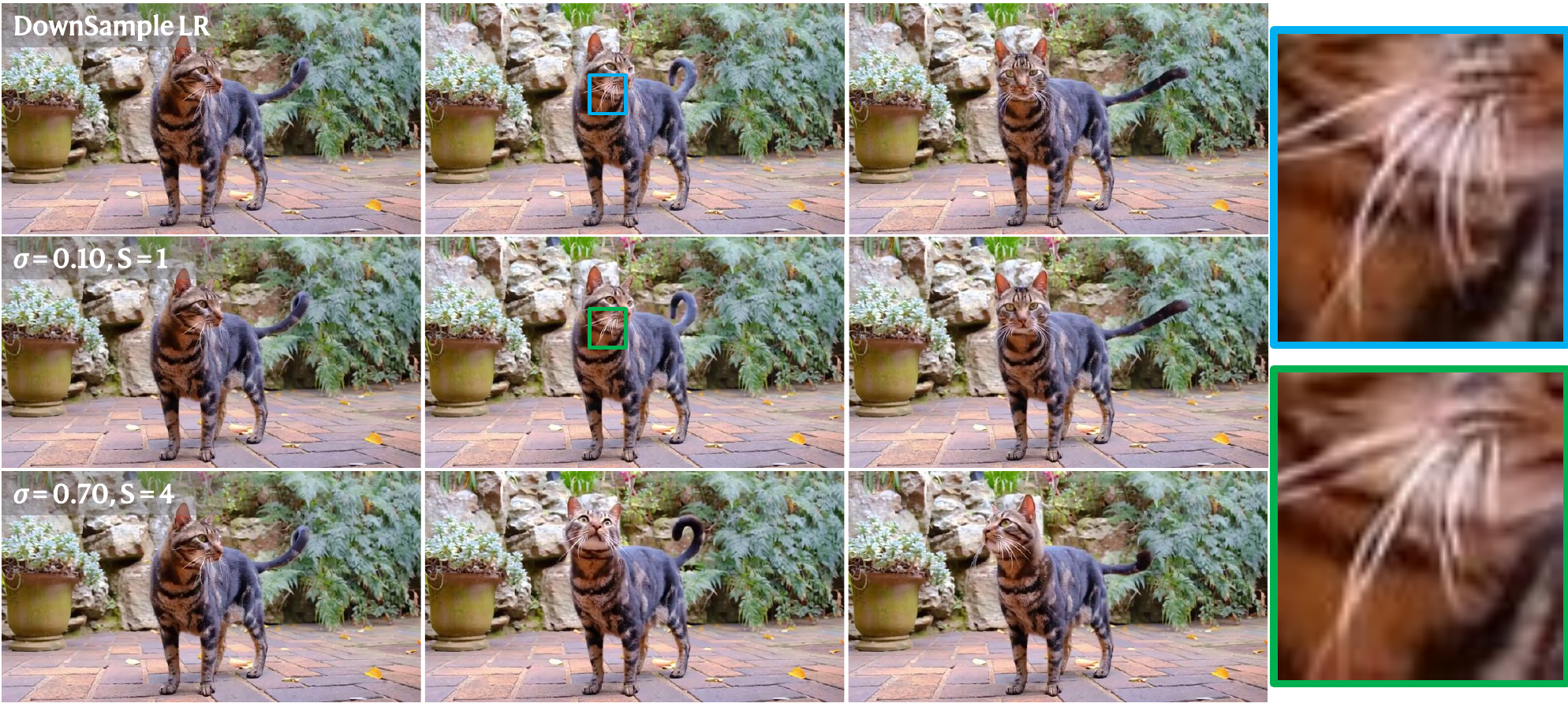}
  \caption{Qualitative comparison of stage-transition synthesis under different settings.}
  \label{fig:stage-trans}
\end{figure}

We run a test on 36 videos and compute SNR/PSNR/SSIM between \(\tilde V^{LR}\) and \(V^{LR}\), shown in Table~\ref{tab:stage-trans}.
When \(\sigma\) is too small (e.g., \(0.01\)), the artifact injection is insufficient; when \(\sigma\) is too large (e.g., \(0.70\)), the perturbation becomes excessive and the motion is corrupted (Figure~\ref{fig:stage-trans}).
We thus choose \(\sigma=0.10\) as a practical trade-off, and adopt \(S=1\) for efficient large-scale data synthesis.

\begin{table}[t]
  \centering
  \small
  \setlength{\tabcolsep}{10pt}
  \caption{Stage transition diagnostics under different noise strengths \(\sigma\) and denoising steps \(S\).}
  \label{tab:stage-trans}
  \begin{tabular}{cccccc}
    \toprule
    \(\sigma\) & \(S\) & SNR \(\uparrow\) & PSNR \(\uparrow\) & SSIM \(\uparrow\) \\
    \midrule
    0.01 & 1 & 7.412 & 35.165 & 0.896 \\
    0.1 & 4 & 5.948 & 33.701 & 0.833 \\
    0.1 & 1 & 6.152 & 33.905 & 0.841 \\
    0.3 & 4 & 5.069 & 32.828 & 0.780 \\
    0.5 & 4 & 4.593 & 32.346 & 0.748 \\
    0.7 & 4 & 4.013 & 31.766 & 0.702 \\
    \bottomrule
  \end{tabular}
\end{table}

\subsection{VAE Reconstruction Fidelity}
\label{app:vae-fidelity}

To quantitatively validate that the higher compression ratio of the VAE used in Stage~II does not compromise the recovery of high-frequency details, we conduct a systematic VAE reconstruction fidelity experiment. Specifically, we extract 93 video clips (each with 81 frames) from 31 open-domain 4K source videos. We perform a full VAE encode-decode cycle on these clips at three different resolutions: 4K ($3840\times 2160$), 2K ($2560\times 1440$), and 720P ($1280\times 720$). We measure pixel-level metrics (PSNR, SSIM) and a perceptual metric (LPIPS). The results are summarized in Table~\ref{tab:vae_reconstruction}.

\begin{table}[H]
\centering
\small
\caption{VAE reconstruction quality at different resolutions.}
\label{tab:vae_reconstruction}
\begin{tabularx}{0.6\textwidth}{lCCC}
\toprule
Resolution & PSNR (dB) $\uparrow$ & SSIM $\uparrow$ & LPIPS $\downarrow$ \\
\midrule
4K ($3840\times 2160$) & 36.94 & 0.955 & 0.051 \\
\textbf{2K ($2560\times 1440$)} & \textbf{35.25} & \textbf{0.941} & \textbf{0.049} \\
720P ($1280\times 720$) & 33.20 & 0.914 & 0.048 \\
\bottomrule
\end{tabularx}
\end{table}

\begin{figure}[H]
    \centering
    \includegraphics[width=0.8\linewidth]{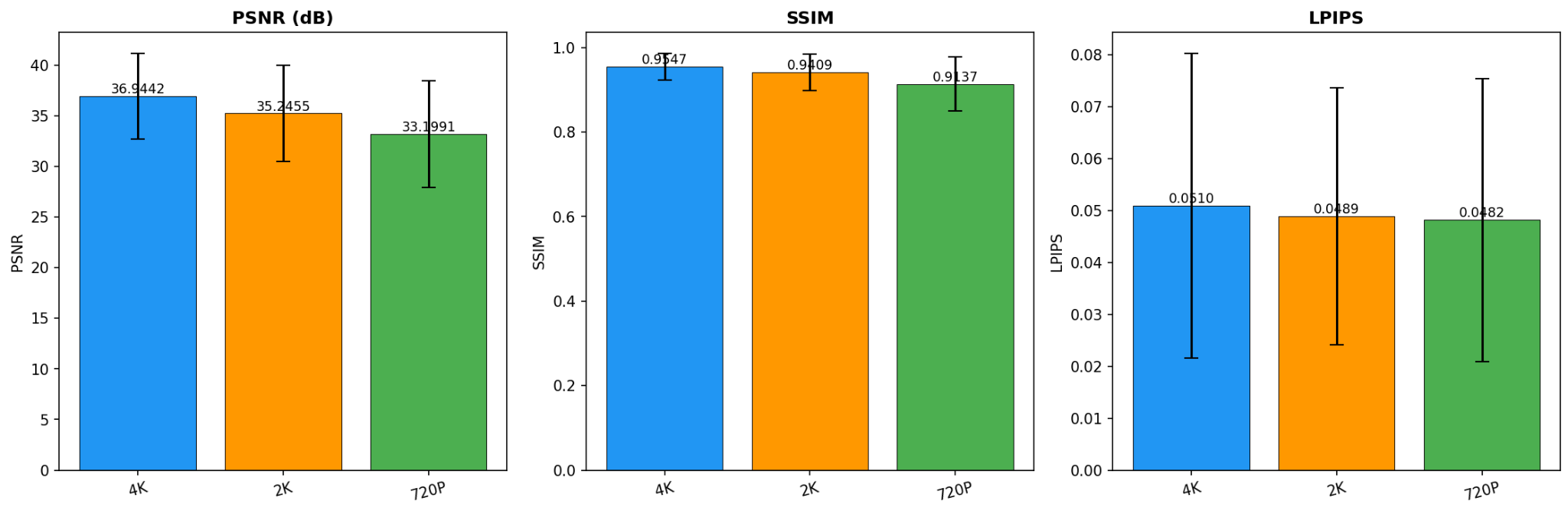}
    \caption{LPIPS comparison of VAE reconstruction across different resolutions.}
    \label{fig:vae_lpips}
\end{figure}

As shown in Table~\ref{tab:vae_reconstruction}, at the target 2K resolution of Stage~II, the VAE achieves excellent reconstruction quality with a PSNR of 35.25 dB and an SSIM of 0.941, indicating high fidelity in preserving video content, including high-frequency details. Furthermore, the perceptual quality remains highly consistent across different resolutions, as evidenced by the similar LPIPS values (Figure~\ref{fig:vae_lpips}).

\begin{figure}[H]
    \centering
    \includegraphics[width=0.6\linewidth]{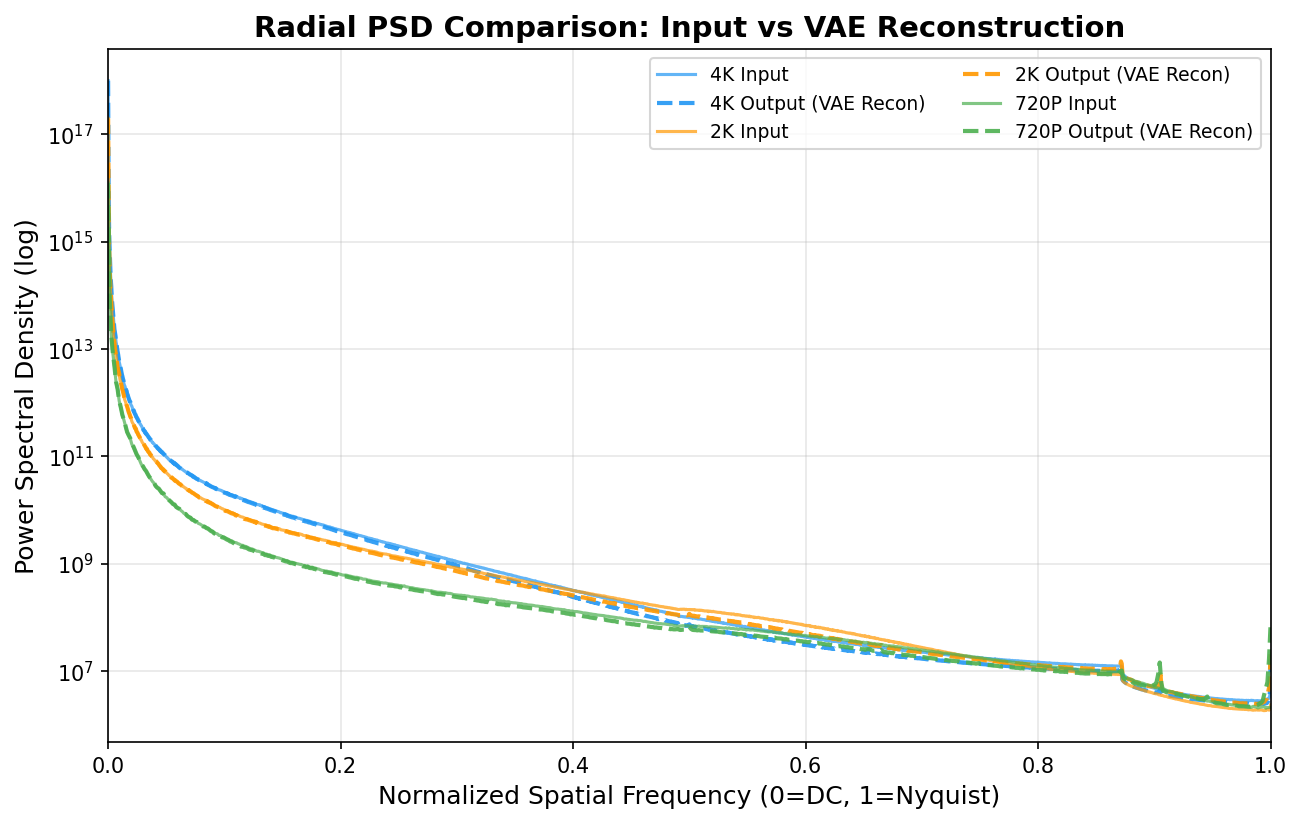}
    \caption{Radial Power Spectral Density (PSD) comparison between input and VAE reconstructed videos.}
    \label{fig:vae_psd}
\end{figure}

To further confirm the preservation of high-frequency information, we compare the radial power spectral density (PSD) of the input videos and the VAE reconstructed videos across different resolutions (Figure~\ref{fig:vae_psd}). The PSD curves of the input and reconstructed videos almost perfectly overlap, even in the high-frequency regions (normalized frequency $> 0.6$). This spectral analysis directly verifies that the VAE faithfully retains full-band information, from low to high frequencies, despite the high compression ratio.

In conclusion, while Stage~II employs a VAE with a higher compression ratio to reduce the token count, the spatial information required per latent token at the target 2K resolution is well within the VAE's representational capacity. This design achieves a favorable balance between quality and efficiency, significantly reducing the computational overhead of Stage~II without sacrificing the fidelity of high-frequency detail recovery.

\subsection{Temporal Smoothness across Segment Boundaries}
\label{app:temporal-smoothness}

To verify that our Conditional Segment-wise Generation (CSG) and bidirectional contextual interaction effectively eliminate segmentation-induced discontinuities, we conduct a quantitative analysis of temporal smoothness across segment boundaries. We measure frame-pair dissimilarity at the 6 CSG segment boundaries versus non-boundary positions across VBench-I2V subset test videos. As a discontinuity-free reference, we include the \textbf{Upscale} baseline (Stage~I + bicubic upsampling), which involves no segmentation.

We evaluate three metrics: Optical Flow, $1-\text{SSIM}$, and Pixel Difference. The results are summarized in Table~\ref{tab:temporal_smoothness}.

\begin{table}[H]
\centering
\small
\caption{Frame-pair dissimilarity at boundary vs. non-boundary positions.}
\label{tab:temporal_smoothness}
\begin{tabularx}{0.9\textwidth}{lCCCCC}
\toprule
Metric & Upscale Bnd & Upscale Non & SwiftI2V Bnd & SwiftI2V Non & $\Delta_{\text{Up}}$ / $\Delta_{\text{Ours}}$ \\
\midrule
Optical Flow & 0.727 & 0.706 & 0.705 & 0.688 & $+2.9\%$ / \textbf{$+2.5\%$} \\
$1-\text{SSIM}$ & 0.115 & 0.109 & 0.133 & 0.131 & $+6.1\%$ / \textbf{$+1.9\%$} \\
Pixel Diff & 4.37 & 4.10 & 4.72 & 4.67 & $+6.4\%$ / \textbf{$+1.1\%$} \\
\bottomrule
\end{tabularx}
\end{table}

Across all three metrics, the boundary-vs-non-boundary gap of SwiftI2V is consistently smaller than that of the segmentation-free reference: $+2.5\%$ vs. $+2.9\%$ for optical flow, $+1.9\%$ vs. $+6.1\%$ for $1-\text{SSIM}$, and $+1.1\%$ vs. $+6.4\%$ for pixel difference. 

\begin{figure}[H]
    \centering
    \includegraphics[width=0.9\linewidth]{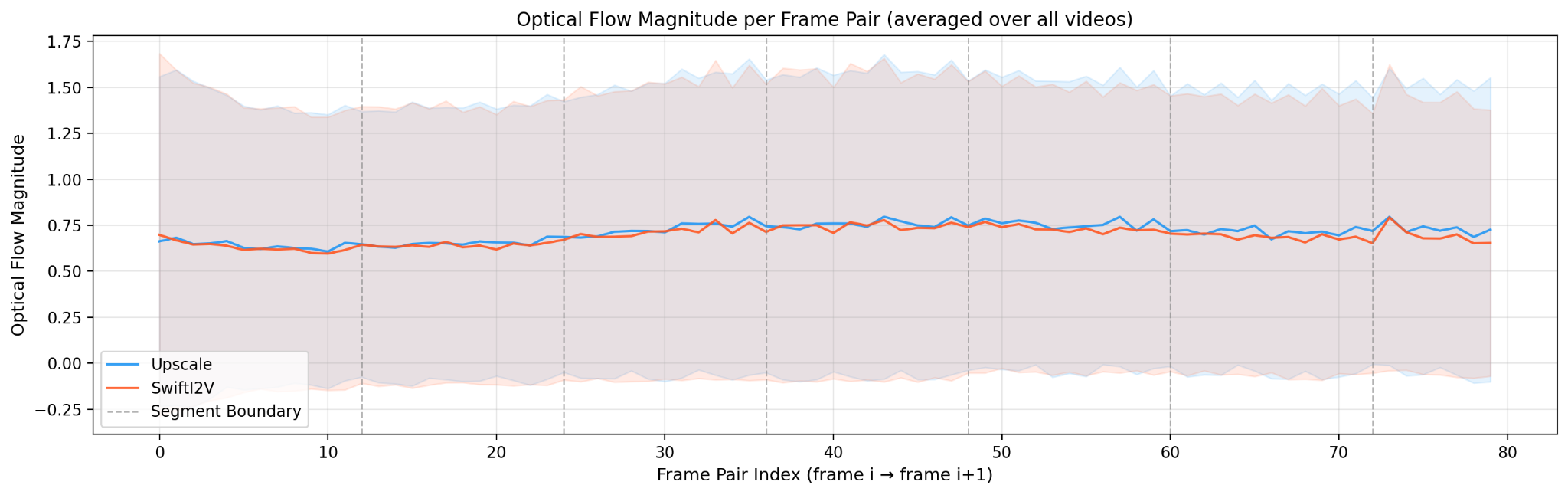}
    \caption{Per-frame optical flow curve across segment boundaries.}
    \label{fig:optical_flow_boundary}
\end{figure}

Furthermore, the per-frame optical flow curve (Figure~\ref{fig:optical_flow_boundary}) confirms the absence of any spikes at boundary positions. These results validate that our CSG design, combined with bidirectional contextual interaction, effectively ensures temporal smoothness and prevents subtle temporal ``popping'' artifacts at segment boundaries.

\subsection{Analysis of Cross-Segment Error Accumulation}
\label{app:error-accumulation}

To comprehensively analyze how our Conditional Segment-wise Generation (CSG) mitigates error accumulation across segments, we conduct a systematic segment-wise quality degradation experiment. We evaluate the frame-level image quality metrics MUSIQ ($\uparrow$) and NIQE ($\downarrow$) independently for each segment on the VBench-I2V subset. We then perform linear regression on the segment scores to quantify the degradation trend. We compare three settings: \textbf{SwiftI2V} (with bidirectional interaction), \textbf{AR} (the \texttt{w/o bi-interaction} variant using a unidirectional causal mask), and \textbf{Upscale} (Stage~I output directly upsampled, serving as a reference baseline without high-resolution detail synthesis).

\begin{figure}[H]
    \centering
    \begin{minipage}{0.48\textwidth}
        \centering
        \includegraphics[width=\linewidth]{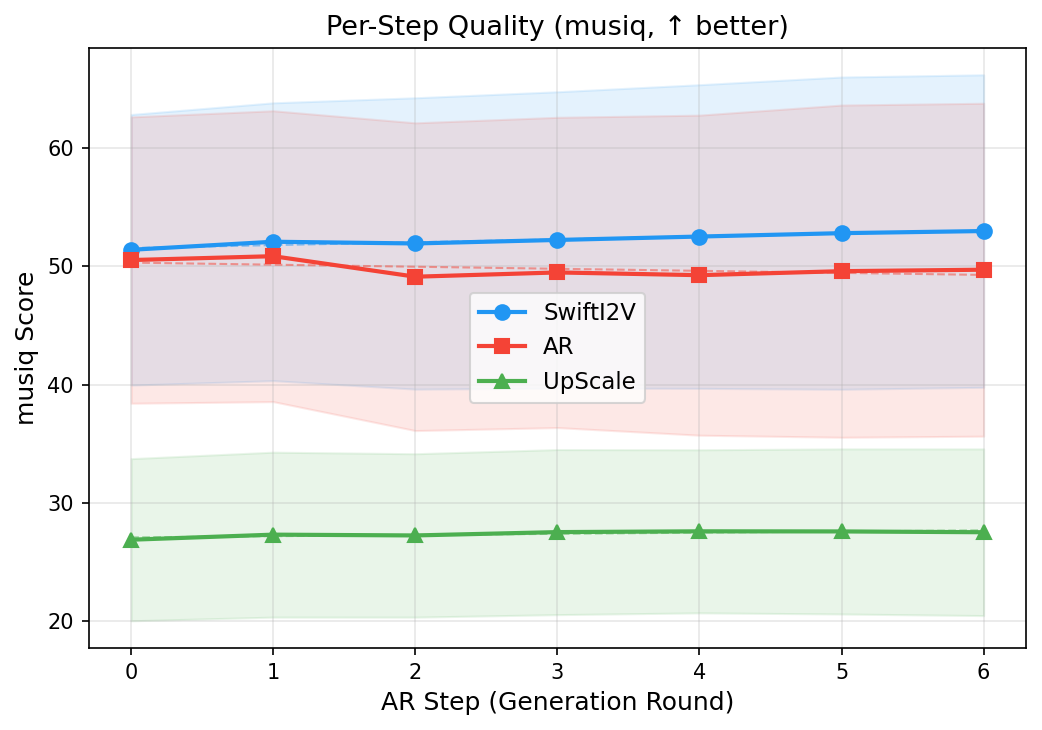}
        \caption{Segment-wise MUSIQ trend.}
        \label{fig:musiq_degradation}
    \end{minipage}\hfill
    \begin{minipage}{0.48\textwidth}
        \centering
        \includegraphics[width=\linewidth]{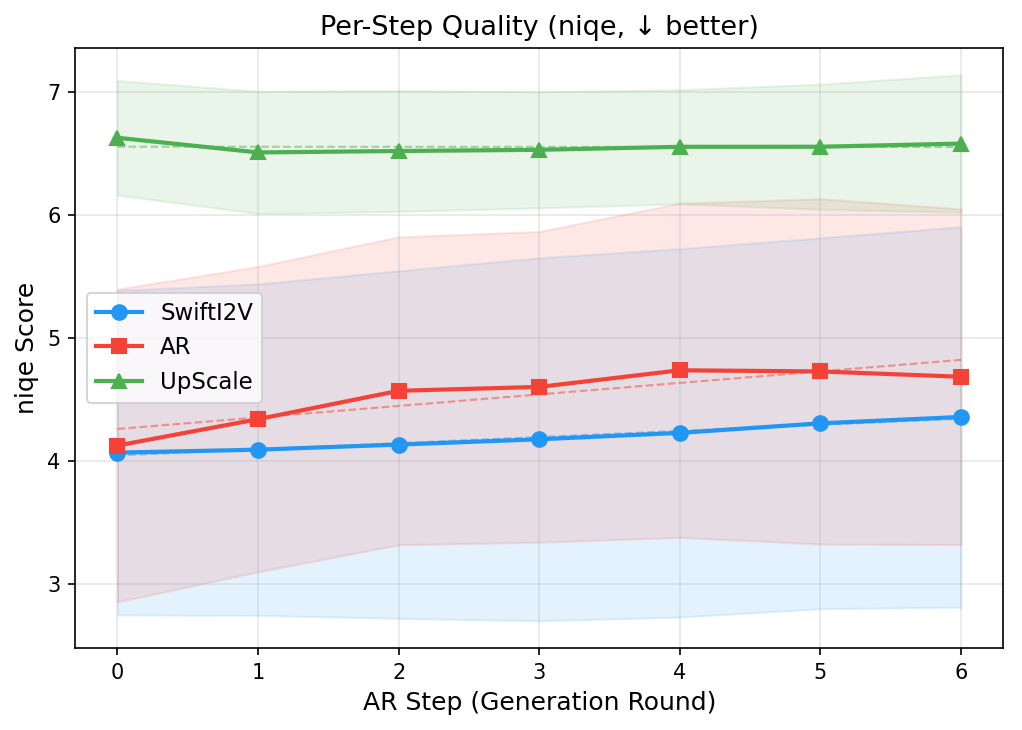}
        \caption{Segment-wise NIQE trend.}
        \label{fig:niqe_degradation}
    \end{minipage}
\end{figure}

The quantitative results are summarized in Table~\ref{tab:error_accumulation}, and the corresponding degradation trends are visualized in Figure~\ref{fig:musiq_degradation} and Figure~\ref{fig:niqe_degradation}.

\begin{table}[H]
\centering
\small
\caption{Segment-wise degradation analysis of MUSIQ and NIQE.}
\label{tab:error_accumulation}
\begin{tabularx}{0.8\textwidth}{lCCC}
\toprule
Method & NIQE Slope $\downarrow$ & NIQE Degradation/step $\downarrow$ & MUSIQ Slope $\uparrow$ \\
\midrule
\textbf{SwiftI2V} & \textbf{$+0.050$} & \textbf{$1.22\%$} & \textbf{$+0.243$} \\
AR & $+0.093$ & $2.26\%$ & $-0.174$ \\
Upscale & $-0.001$ & $0.01\%$ & $+0.100$ \\
\bottomrule
\end{tabularx}
\end{table}

As shown in Table~\ref{tab:error_accumulation} and Figure~\ref{fig:niqe_degradation}, the AR variant exhibits a clear error accumulation trend in NIQE, with a degradation slope of $+0.093$/step (a $2.26\%$ degradation per step). In contrast, SwiftI2V significantly suppresses this accumulation, reducing the slope to $+0.050$/step (a $1.22\%$ degradation per step), which is only $53\%$ of the AR degradation rate. Furthermore, the absolute NIQE scores of SwiftI2V remain consistently better than those of AR across all segments.

For MUSIQ (Figure~\ref{fig:musiq_degradation}), the AR variant shows a downward trend (slope $-0.174$), indicating a loss of detail quality over time. Conversely, SwiftI2V maintains a stable and even slightly improving trend (slope $+0.243$), demonstrating that the detail synthesis quality is well-preserved in later segments.

These results confirm that the bidirectional contextual interaction mechanism effectively mitigates the cascading error accumulation commonly observed in autoregressive generation. By allowing conditioning blocks to actively participate in the attention computation and dynamically adapt to the current segment's denoising needs, SwiftI2V prevents the propagation of imperfect historical information, thereby maintaining high fidelity throughout the progressive generation process.

\subsection{Analysis of Dynamic Degree}
\label{app:dynamic-degree}

In Table~\ref{tab:result}, SwiftI2V achieves a Dynamic Degree score of 0.3008. While this score is slightly lower than that of Stream-DiffVSR (0.3374), it is important to interpret these metrics in the context of visual quality and the two-stage generation paradigm.

First, compared to direct end-to-end generation models such as LTX-2 (0.0488) and CineScale (0.1667), SwiftI2V demonstrates a significantly higher Dynamic Degree. This indicates that our method is highly capable of generating substantial and meaningful motion, overcoming the common issue of static outputs in high-resolution I2V generation.

Second, the unusually high Dynamic Degree score of Stream-DiffVSR is largely an artifact of temporal flickering and structural instability, rather than coherent semantic motion. As observed in qualitative comparisons, Stream-DiffVSR struggles to maintain the structural integrity of the input image at 2K resolution, leading to unnatural high-frequency temporal variations that artificially inflate the Dynamic Degree metric.

Finally, to address the concern that Stage~II might overly constrain motion to preserve spatial details, we compare the Dynamic Degree of SwiftI2V with its Stage~I output (denoted as \textcolor[gray]{0.35}{Upscale} in Table~\ref{tab:result}). The Upscale baseline achieves a Dynamic Degree of 0.2805. After the Stage~II refinement, the score actually increases to 0.3008. This quantitative evidence demonstrates that Stage~II not only faithfully preserves the motion dynamics established by Stage~I but also further enhances them, confirming that our detail-preserving mechanisms do not come at the expense of motion magnitude.

\section{More 2K I2V Visual Results of SwiftI2V}
\label{app:visual}

\begin{figure*}[t]
    \centering
    \includegraphics[width=0.93\linewidth]{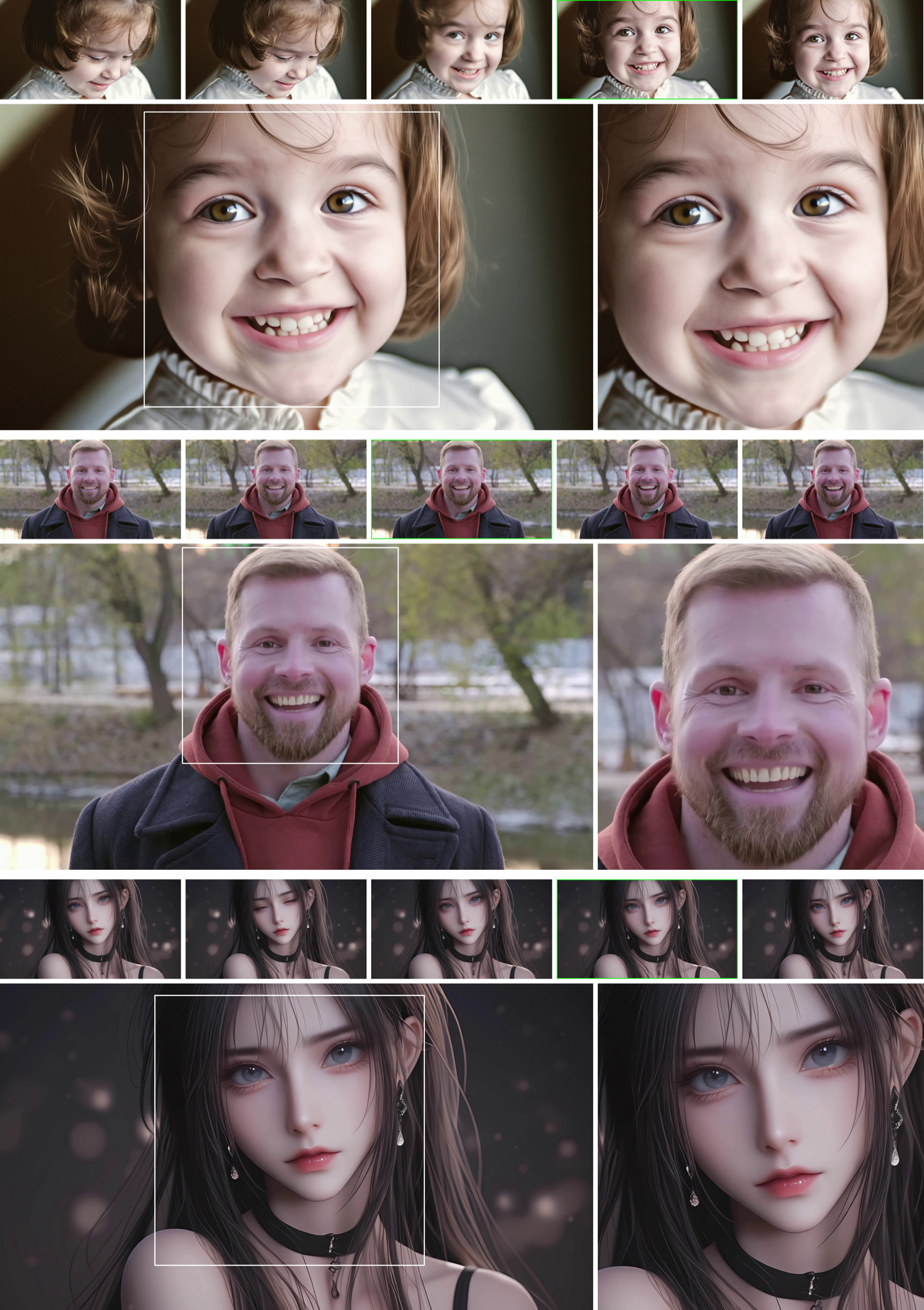}
    \caption{More 2K I2V visual results of SwiftI2V.}
    \label{fig:more_result_1}
\end{figure*}

\begin{figure*}[t]
    \centering
    \includegraphics[width=0.93\linewidth]{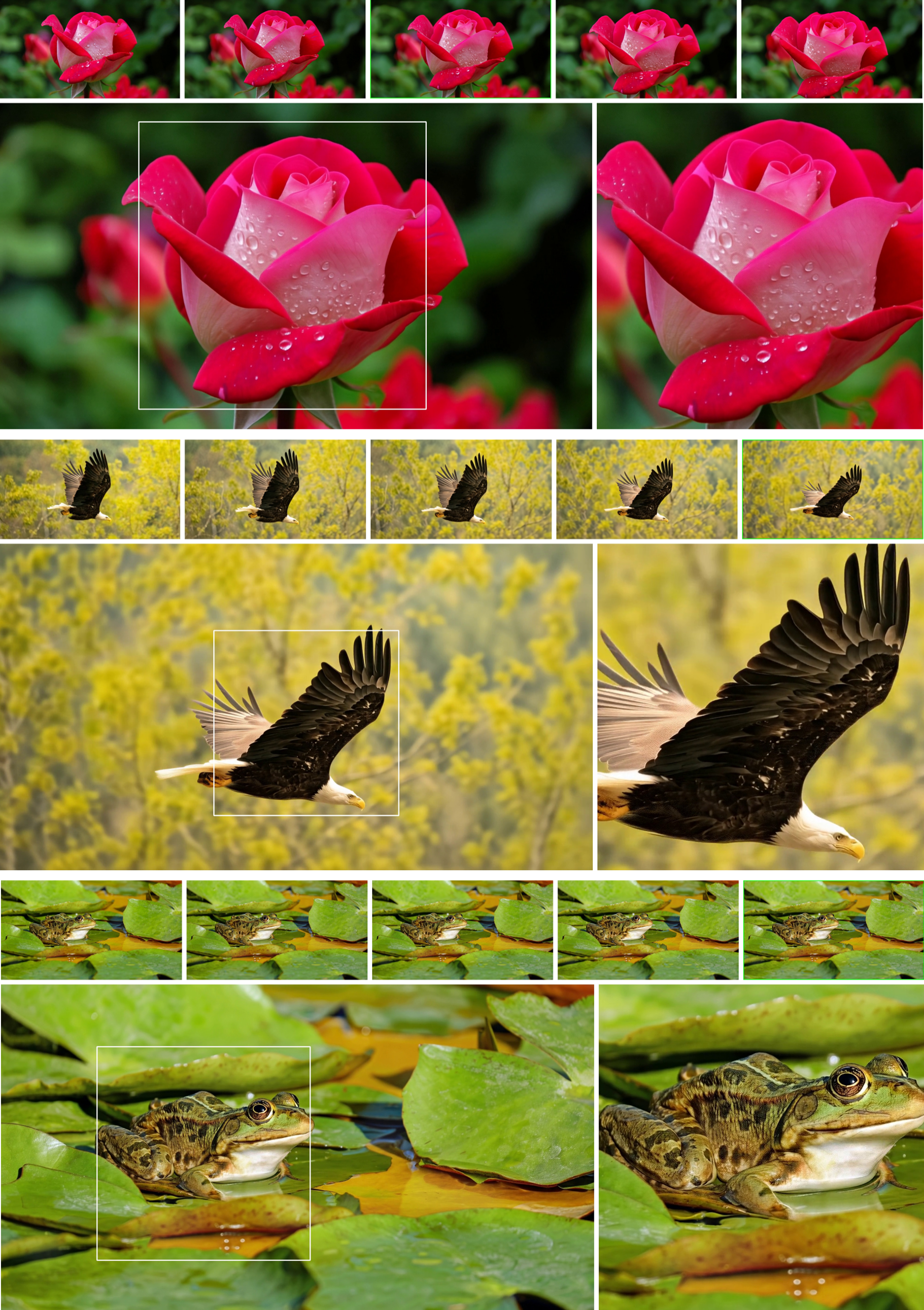}
    \caption{More 2K I2V visual results of SwiftI2V.}
    \label{fig:more_result_2}
\end{figure*}

\begin{figure*}[t]
    \centering
    \includegraphics[width=0.93\linewidth]{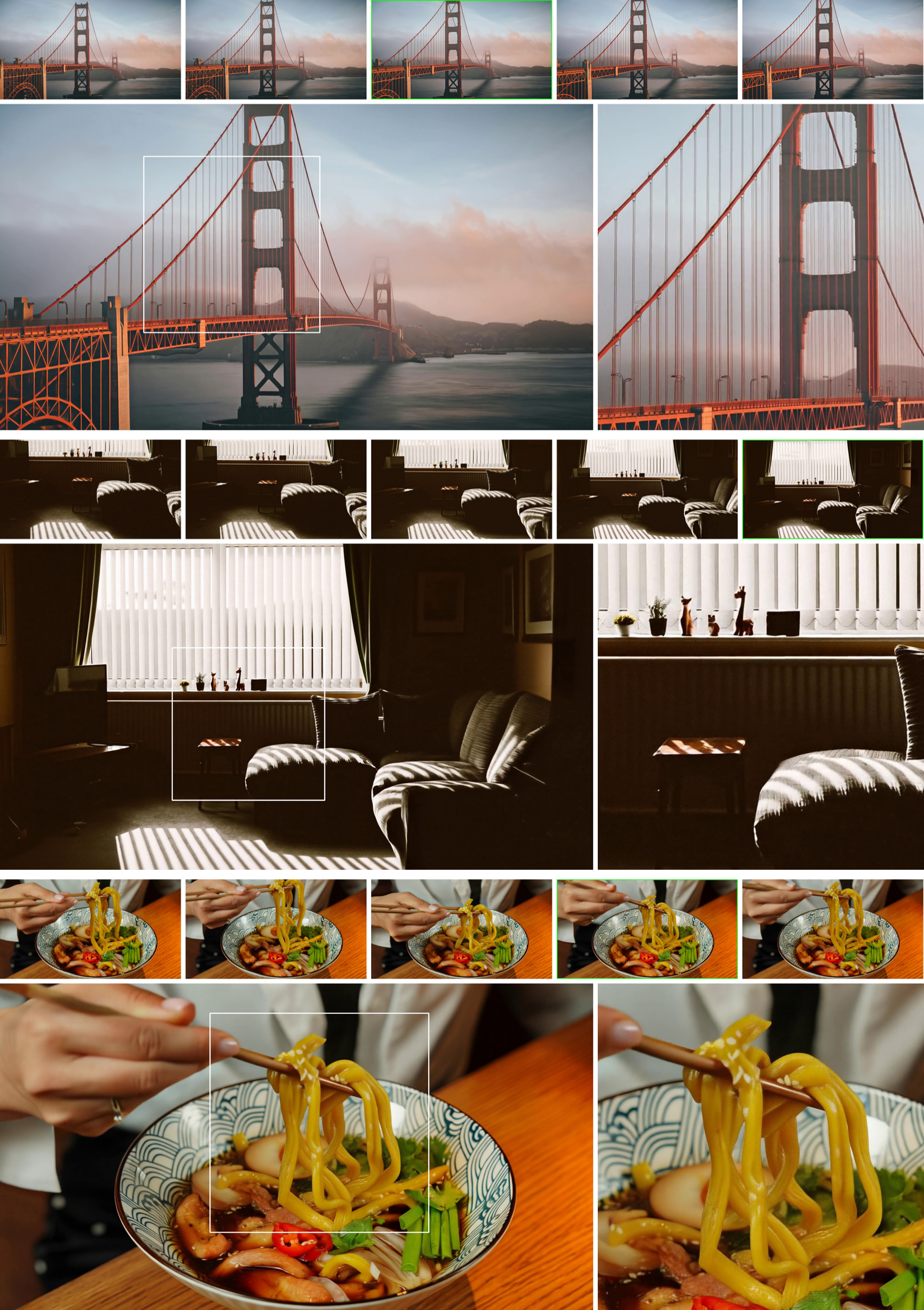}
    \caption{More 2K I2V visual results of SwiftI2V.}
    \label{fig:more_result_3}
\end{figure*}

We provide additional 2K image-to-video generation results produced by SwiftI2V in Figures~\ref{fig:more_result_1}, \ref{fig:more_result_2}, and \ref{fig:more_result_3}. These examples demonstrate that SwiftI2V can synthesize temporally coherent dynamics while faithfully preserving input-specific spatial structures and fine-grained textures under strong image conditioning.

\clearpage

\clearpage

\clearpage

\end{document}